\documentclass[11pt]{article}

\usepackage[preprint]{acl}

\usepackage{times}
\usepackage{latexsym}
\usepackage[T1]{fontenc}
\usepackage[utf8]{inputenc}
\usepackage{microtype}
\usepackage{inconsolata}
\usepackage{graphicx}
\usepackage{amsmath}
\usepackage{amssymb}
\usepackage{booktabs}
\usepackage{array}
\usepackage{tabularx}
\usepackage{multirow}
\usepackage[table]{xcolor}
\usepackage{longtable}

\newcolumntype{L}{>{\raggedright\arraybackslash}X}
\newcolumntype{P}[1]{>{\raggedright\arraybackslash}p{#1}}

\definecolor{ARCHERYellow}{RGB}{255,249,196}

\title{ARCHER: From Rationale Imitation to Executable Reasoning in Small Language Models}

\author{
Dhruv Dixit$^{1}$,
Khushi Patel$^{2}$,
Congcong Xu$^{3}$,
Min Song$^{4}$ \\[2pt]
$^{1,3,4}$Department of Electrical and Computer Engineering \\
$^{2}$Department of Computer Science \\
Stevens Institute of Technology, Hoboken, NJ - 07030, United States \\
$^{1}$ddixit1@stevens.edu, $^{2}$kpatel105@stevens.edu, \\
$^{3}$cxu7@stevens.edu, $^{4}$msong6@stevens.edu
}

\begin{document}

\maketitle

\footnotetext{Our code is available at this anonymous \href{https://anonymous.4open.science/r/ARCHER_HiGRA_TRACE-B609}{ link}.}

\begin{abstract}
Small language models are attractive for local, low-cost, and privacy-sensitive deployment, but they remain unreliable on mathematical reasoning tasks that require decomposition, exact computation, tool use, and verification. We introduce \textbf{ARCHER}: \textbf{A}ction-trace \textbf{R}einforcement for \textbf{C}alculator-grounded \textbf{H}ierarchical \textbf{E}xecutable \textbf{R}easoning, a two-stage framework for training small language models (SLMs) to produce executable and auditable reasoning trajectories. In the first stage, ARCHER compiles free-form chain-of-thought rationales into typed action traces containing reasoning steps, calculator calls, deterministic runtime observations, verification actions, and final answers. This shifts the supervision target from imitating explanatory text to learning a tool-grounded reasoning policy. In the second stage, ARCHER refines this policy through asymmetric hierarchical reinforcement learning, using grouped and conditionally gated trajectory rewards for answer correctness, tool grounding, process discipline, and trace alignment. Across five standard sub-10B backbones, ARCHER improves over the strongest tool-augmented baseline by 5.51 percentage points on macro-average across benchmark accuracies, including gains of 8.95 points on MATH-500 and 5.33 and 5.34 points on AIME-2024 and AIME-2025. Evaluations on two additional reasoning-specialized 7B backbones further show that executable grounding remains beneficial after reasoning-focused reinforcement learning or distillation. These results demonstrate that executable supervision and reward-guided behavioral refinement provide a general and effective training paradigm for compact mathematical reasoners. 
\end{abstract}

\section{Introduction}

Small language models (SLMs) are attractive for local, low-cost, and privacy-sensitive deployment, but remain unreliable on mathematical tasks requiring decomposition, exact computation, and verification. A correct solution must identify relevant quantities, select operations, preserve intermediate results, and produce a well-formed answer. This motivates training SLMs to generate reasoning trajectories that can be executed and checked, rather than merely fluent explanations.

Chain-of-thought prompting established the value of intermediate reasoning \citep{wei2022cot}, motivating rationale supervision and reasoning distillation for smaller models \citep{hsieh2023distilling,shridhar2023distilling,gu2023minillm,ranaldi2024aligning}. Prior work also studies rationale faithfulness \citep{wang-etal-2023-scott}, efficient reasoning transfer \citep{chen2025skipthinking}, and small-large model collaboration \citep{zhao2025dre}. However, free-form rationales are non-canonical and often mix reasoning decisions with implicit arithmetic. A compact student may therefore imitate the form of a solution without learning which computations must be executed or how their results should guide later steps.

Tool-augmented methods address this problem by externalizing computation. Program-aided approaches separate language reasoning from executable operations \citep{gao2023pal,chen2023program}, while faithful reasoning methods connect generated explanations to executable derivations \citep{lyu2023faithful}. Other systems interleave reasoning with tools or APIs \citep{yao2023react,schick2023toolformer,lu2024gear}, including calculator-oriented methods for mathematical reasoning \citep{kadlvcik2023calc,chen2024good}. These approaches demonstrate the value of execution, but tool access alone does not teach a compact model when to invoke a tool, how to construct a valid call, or how to use the returned observation consistently.

Process supervision and reinforcement learning provide complementary mechanisms for improving reasoning behavior. Process-level methods evaluate intermediate steps rather than only final answers \citep{lightman2023let,wang2024math}, while recent reasoning models use verifiable outcomes and structured reinforcement-learning objectives \citep{guo2025deepseek,team2025kimi}. Tool-integrated reinforcement learning further shows that execution feedback can improve interaction policies \citep{feng2025retool}. Nevertheless, reward design remains difficult: sparse outcome signals provide weak credit assignment, while process rewards can be brittle, noisy, or misaligned with reasoning-model outputs \citep{dai2025stable,dontsov2026out}.

In this paper, we address the intersection of these three lines of work. Unlike rationale distillation, our framework replaces free-form explanations with typed executable actions. Unlike tool-access methods, it makes calculator calls, runtime observations, verification, and termination part of the supervised trajectory. Unlike outcome-only or flat process-reward optimization, it organizes feedback into priority-ordered reward families that are activated only when relevant.

We propose \textbf{ARCHER}: \textbf{A}ction-trace \textbf{R}einforcement for \textbf{C}alculator-grounded \textbf{H}ierarchical \textbf{E}xecutable \textbf{R}easoning, a two-stage framework for training SLMs as executable mathematical reasoning policies. In Stage~1, \textbf{TRACE} (\textbf{T}ool-augmented \textbf{R}easoning via \textbf{A}ction-\textbf{C}ompiled \textbf{E}xecution) converts legacy chain-of-thought rationales into typed trajectories containing reasoning actions, calculator calls, deterministic observations, verification actions, and final answers. In Stage~2, \textbf{HiGRA} (\textbf{Hi}erarchically \textbf{G}ated \textbf{R}eward \textbf{A}ggregation) refines the TRACE policy using priority-ordered, conditionally gated rewards for answer correctness, tool grounding, process discipline, and trace alignment. We evaluate ARCHER across seven sub-10B backbones and eight mathematical reasoning benchmarks. Across five standard backbones, ARCHER improves over the strongest tool-augmented baseline by 5.51 percentage points on average and is best or tied-best in all 40 model-benchmark comparisons; it also improves both reasoning-specialized backbones across all eight benchmarks.

\paragraph{Contributions.}
ARCHER makes three contributions. First, it formulates SLM mathematical reasoning as executable policy learning, with typed model actions and deterministic external computation. Second, a compiler framework that converts existing rationale datasets into calculator-grounded supervision without requiring manually authored action traces. Third, a reinforcement learning algorithm that introduces priority-ordered and conditionally gated trajectory rewards that jointly optimize correctness, tool grounding, process discipline, and trace compatibility. Together, these 3 contributions unify rationale supervision, tool-augmented execution, and process-aware reinforcement learning within a single framework for accurate and auditable mathematical reasoning.


\section{Related Work}

\paragraph{Reasoning supervision and distillation.}
Chain-of-thought prompting established the value of intermediate reasoning \cite{wei2022cot}, motivating rationale supervision and stepwise distillation for smaller models \cite{hsieh2023distilling,shridhar2023distilling}. Subsequent work improves rationale faithfulness \cite{wang-etal-2023-scott}, generative distillation objectives \cite{gu2023minillm}, transfer efficiency \cite{chen2025skipthinking}, and small-large model collaboration \cite{zhao2025dre}. ARCHER instead supervises typed executable actions, delegating computation to a deterministic runtime rather than imitating free-form explanations.

\paragraph{Tool-grounded reasoning.}
Program-aided methods separate language reasoning from executable computation \cite{gao2023pal,chen2023program}, while faithful reasoning connects explanations to executable derivations \cite{lyu2023faithful}. General tool-use frameworks interleave reasoning with actions or APIs \cite{yao2023react,schick2023toolformer}, and calculator-centered methods improve mathematical reasoning through explicit computation and tool feedback \cite{kadlvcik2023calc,chen2024good}. ARCHER makes this interaction the training target itself by compiling legacy rationales into canonical action-observation traces and optimizing their correctness, grounding, and validity.

\paragraph{Process supervision and reinforcement learning.}
Process supervision evaluates intermediate reasoning rather than final answers alone \cite{lightman2023let,wang2024math}, while verifiable RL objectives induce stronger reasoning behavior \cite{guo2025deepseek,team2025kimi}. Tool-integrated and stability-focused studies further expose the difficulty of sparse, noisy, or fragmented rewards \cite{feng2025retool,dai2025stable}. Recent RL-specialized systems include DeepSeek-R1 \citep{guo2025deepseek}, Kimi k1.5 \citep{team2025kimi}, MiMo-7B-RL, and Falcon-H1R \citep{xiaomi2025mimo,falcon2026h1r}. ARCHER complements these text-only reasoners with deterministic execution and differs from flat reward mixtures by using priority-ordered, conditionally gated reward families for correctness, tool grounding, process discipline, and trace alignment.

\section{Methodology}
\label{sec:methodology}

\begin{figure*}[t]
    \centering
    \includegraphics[width=\textwidth]{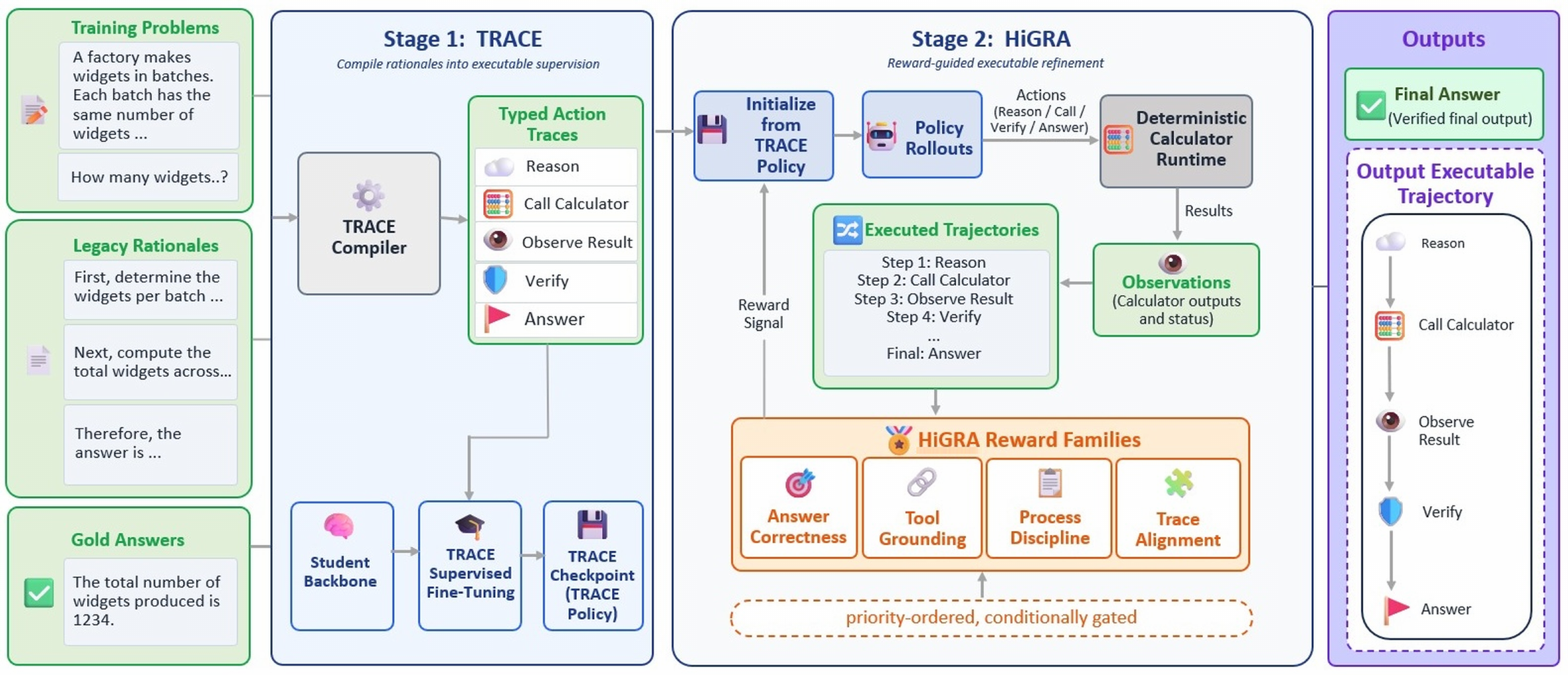}
    \caption{Overview of ARCHER. Modules in Blue are trainable components, Light Gray are non-trainable or deterministic, Orange are Reward or evaluation, and Green are Data/traces/observations. }
    \label{fig:archer_overview}
\end{figure*}

The ARCHER framework, illustrated in Figure~\ref{fig:archer_overview}, has two stages. Stage 1, TRACE, compiles legacy chain-of-thought rationales into calculator-grounded action traces and trains the student by supervised fine-tuning. Stage 2, HiGRA, refines the TRACE policy with grouped and conditionally gated trajectory rewards. TRACE teaches the model the executable action language; HiGRA improves how reliably the model uses that language to solve problems. Working together, ARCHER does not merely expose the model to tools or rationales; it converts rationale supervision into an executable control interface and then optimizes the model over completed, runtime-executed trajectories.

\subsection{Overall Reasoning Policy}

Each training example is a tuple $(x,r,y^\star)$, where $x$ is the input problem, $r$ is a legacy rationale, and $y^\star$ is the gold answer. ARCHER trains a policy $\pi_\theta$ that maps $x$ to an executable trajectory $\tau$ and a final answer $\hat{y}$. A trajectory is a sequence of states $\tau=(s_1,\ldots,s_T)$ and $s_t=(a_t,u_t,o_t)$, where $a_t$ is an action type, $u_t$ is a model-emitted payload, and $o_t$ is an optional runtime observation. We use a compact action vocabulary:
\[
\mathcal{A} =
\left\{
\begin{aligned}
&\texttt{READ},\ \texttt{PLAN},\ \texttt{CALL\_CALC},\\
&\texttt{OBSERVE},\ \texttt{VERIFY},\ \texttt{ANSWER}
\end{aligned}
\right\}
\]
The model generates reasoning actions and calculator-call payloads. Calculator observations are produced by a deterministic runtime rather than hallucinated by the model. The final prediction is extracted from the terminal answer action, $\hat{y}=\mathrm{Ans}(\tau)$.

\subsection{Stage 1: TRACE}

TRACE converts free-form rationales into executable action traces. The compiler $\mathcal{C}$ maps an example $(x,r,y^\star)$ to a gold executable trajectory $\tau^\star=\mathcal{C}(x,r,y^\star)$. The compiler separates textual reasoning spans from arithmetic spans. Textual spans are mapped to actions such as \texttt{READ}, \texttt{PLAN}, and \texttt{VERIFY}. Arithmetic spans are normalized into calculator expressions and emitted as paired tool actions:
\[
\texttt{CALL\_CALC}(e_t)
\rightarrow
\texttt{OBSERVE}(f_{\mathrm{calc}}(e_t)),
\]
where $f_{\mathrm{calc}}$ is the calculator runtime. For GSM-8K, the compiler uses the dataset's inline calculator annotations to identify arithmetic spans, executes each expression, and inserts the resulting observation into the trace. The trace is always terminated with a normalized gold answer. This metadata is not serialized into the model-visible prefix before \texttt{ANSWER}. Gold-answer metadata is neither visible when predicting intermediate actions, nor before the final answer token. It is removed during HiGRA rollout and evaluation and no verification step can directly copy the gold answer.

The student is then trained by supervised fine-tuning on compiled traces:
\[
\begin{gathered}
\mathcal{L}_{\mathrm{TRACE}}(\theta) = \\[-2pt]
-\sum_{t \in \mathcal{M}(\tau^\star)}
\log \pi_\theta
\bigl(a_t^\star,u_t^\star \mid x,s_{<t}^\star\bigr).
\end{gathered}
\]
where $\mathcal{M}(\tau^\star)$ denotes model-generated positions. Runtime observations are included in the context but are not treated as free-form tokens the model must hallucinate. This teaches the student to control the reasoning process while leaving exact arithmetic to the calculator runtime.

\subsection{Stage 2: HiGRA}

TRACE gives the student a structured executable interface, but supervised imitation alone can still produce missing tool calls, wrong calculator inputs, malformed traces, ignored observations, or premature answers. HiGRA addresses these failures by refining the TRACE policy using trajectory-level reinforcement learning. The goal is not only to increase final-answer accuracy, but also to improve the reliability of the executable reasoning process that leads to the answer.

A central design choice in HiGRA is its asymmetric distributed hierarchical reward structure i.e., a priority-weighted grouped reward with conditional tool scoring and gated trace alignment. Rather than treating all feedback signals as a flat collection of equally weighted micro-rewards, HiGRA separates trajectory quality into reward families with different responsibilities and priorities. Outcome correctness remains the dominant signal because the final answer must be correct. Tool grounding is the next most important signal because mathematical reasoning often fails when the model omits a needed calculator call, submits the wrong expression, or ignores the returned observation. Process discipline combines structural-format quality with separate signals for verification, brevity, final-answer presence, and premature termination. Trace alignment is used as the weakest signal: it regularizes the rollout toward the compiled TRACE structure without forcing exact imitation of the supervised trace.

This asymmetry is important because not every reward component should influence every trajectory in the same way. For example, alignment is useful only when the trajectory is already correct or sufficiently tool-grounded; otherwise, rewarding similarity to the compiled trace can reinforce superficial structure even when the computation is wrong. Similarly, tool-use feedback is conditioned on whether the compiled reference trace requires calculator use: required calls are rewarded, while unnecessary calls are discouraged. HiGRA therefore gates auxiliary rewards conditionally, so each reward family contributes only when it is relevant to the current trajectory. This makes HiGRA different from ordinary outcome-only reinforcement learning and from flat process-reward mixtures: it distributes credit across correctness, tool use, process quality, and trace compatibility, but applies these signals asymmetrically according to their reliability and importance.

For a sampled trajectory $\tau$, HiGRA computes a grouped reward:
\begin{flalign}
R_{\mathrm{HiGRA}}(\tau)
&= w_{\mathrm{out}}R_{\mathrm{out}}(\tau)
 + w_{\mathrm{tool}}R_{\mathrm{tool}}(\tau) \nonumber\\
&\quad
 + w_{\mathrm{proc}}R_{\mathrm{proc}}(\tau)
 + w_{\mathrm{align}}\widetilde{R}_{\mathrm{align}}(\tau) \nonumber\\
&\quad
 - P_{\mathrm{len}}(\tau). &&
\label{eq:higra_total}
\end{flalign}

Here, $R_{\mathrm{HiGRA}}(\tau)$ denotes the total reward assigned to trajectory $\tau$. The outcome reward, $R_{\mathrm{out}}(\tau)$, measures final-answer correctness, and $w_{\mathrm{out}}$ denotes its scalar weight. The tool-grounding reward, $R_{\mathrm{tool}}(\tau)$, is conditioned on whether the compiled reference trace requires calculator use and measures call presence, observation pairing, and consistency with deterministic execution; $w_{\mathrm{tool}}$ is its corresponding weight. Malformed or non-executable calls are still counted as generated calls but receive no consistency credit. The process-discipline reward, $R_{\mathrm{proc}}(\tau)$, combines a fractional structural-format score with separate indicators for verification presence, brevity, final-answer presence, and premature termination, while $w_{\mathrm{proc}}$ controls its contribution. The gated alignment reward, $\widetilde{R}_{\mathrm{align}}(\tau)$, softly regularizes the rollout toward the compiled TRACE structure and is weighted by $w_{\mathrm{align}}$; the tilde indicates that the underlying alignment reward is conditionally gated before being added to the total reward. Finally, $P_{\mathrm{len}}(\tau)$ denotes the trajectory-length penalty, which discourages repetitive or overly long trajectories. Full operational definitions of the reward families, weights, gates, reported scores, and penalty are provided in Appendix~\ref{app:archer_details}.

We optimize the TRACE-initialized policy with Group Relative Policy Optimization (GRPO) Reinforcement Learning. For each prompt, we sample a group of trajectories, execute tool calls through the calculator runtime, score each trajectory with $R_{\mathrm{HiGRA}}$, and update the policy using group-relative advantages. This allows ARCHER to optimize non-differentiable trajectory properties such as answer extraction, calculator consistency, schema validity, and trace alignment without training a separate value model.

\subsection{Inference}

At inference time, ARCHER runs the model in the same executable loop used during training. The model emits one action at a time. Non-tool actions are appended directly to the transcript. When the model emits \texttt{CALL\_CALC}, the calculator executes the expression and appends an \texttt{OBSERVE} action. The process stops when the model emits \texttt{ANSWER} or reaches a step budget. Thus, the model is evaluated as a calculator-grounded executable reasoning policy rather than as a free-form text generator.

Full implementation details, reward definitions, and the GRPO objective are provided in Appendix~\ref{app:archer_details}.

\subsection{Why ARCHER Uses Two Stages}

TRACE and HiGRA play complementary roles. TRACE converts raw rationale supervision into an executable action language and gives the compact model an initial tool-using policy. HiGRA then improves that policy with rewards that evaluate complete executed trajectories. This separation is important because reward optimization from an unstructured policy is unstable, while supervised imitation alone can overfit to compiled traces and leave tool-use failures unresolved. ARCHER combines both: TRACE teaches the model how to produce executable reasoning traces, and HiGRA improves whether those traces solve the problem reliably.

\section{Experimental Setup}
\label{sec:experimental_setup}

\paragraph{Models and training.}
We evaluate ARCHER on seven sub-10B backbones: \texttt{Qwen3-1.7B} and \texttt{Qwen3-8B} with thinking mode enabled \citep{yang2025qwen3}, \texttt{Gemma-4-E2B} \citep{gemmateam2026gemma4}, \texttt{Qwen3.5-4B} \citep{qwen2026qwen35}, \texttt{Falcon-H1R-7B} \citep{falcon2026h1r}, \texttt{DeepSeek-R1-Distill-Qwen-7B} \citep{guo2025deepseek}, and \texttt{MiMo-7B-RL} \citep{xiaomi2025mimo}. The first five models form the main controlled comparison, while DeepSeek-R1-Distill-Qwen-7B and MiMo-7B-RL test whether executable grounding complements reasoning-specialized initialization. On \texttt{Falcon-H1R-7B}, used for stage-wise analysis and ablations, we compare the released checkpoint, TRACE-only \texttt{ARCHER w/o HiGRA}, and full \texttt{ARCHER}. For the reasoning-specialized backbones, \texttt{Vanilla}, \texttt{CoA}, and \texttt{ARCHER} denote the released checkpoint, tool-augmented baseline, and complete TRACE-HiGRA pipeline, respectively. Training data, rewards, generation budgets, decoding, and evaluation are otherwise fixed; architecture-specific adaptations are described in Appendix~\ref{app:implementation_details}.

\paragraph{Training data and runtime.}
TRACE compiles each example $(x,r,y^\star)$ into a typed trajectory containing reasoning actions, calculator calls, deterministic observations, verification, and a terminal answer. For GSM-8K, using the official 7,473/1,319 train/test splits, inline calculator annotations identify arithmetic spans; expressions are normalized, executed, and inserted as \texttt{OBSERVE} actions. Runtime outputs condition subsequent generation but are not model-generated decisions. The same deterministic runtime executes \texttt{CALL\_CALC} actions during HiGRA rollouts and evaluation. GSM-8K is used for all stage-wise comparisons, ablations, and trajectory diagnostics because its annotated solutions support evaluation of tool necessity, call construction, observation consistency, trace validity, and process structure; the remaining benchmarks measure answer-level transfer.

\paragraph{Benchmarks.}
GSM-8K is the primary in-domain benchmark \citep{cobbe2021training}. We evaluate transfer to SVAMP \citep{patel2021nlp}, the mathematical subset of MMLU (MM) \citep{hendrycks2020measuring}, MATH \citep{hendrycks2021measuring}, and its representative MATH-500 subset \cite{lightman2024lets}. We further evaluate on the competition-level AIME-2024 and AIME-2025 sets \citep{balunovic2025matharena}, and on multilingual PolyMATH \citep{wang2025PolyMATH}.

\begin{table*}[t]
\centering
\small
\setlength{\tabcolsep}{1mm}
\resizebox{\textwidth}{!}{%
\begin{tabular}{@{}llrrrrrrrr@{}}
\toprule
\textbf{Backbone}
& \textbf{Method}
& \textbf{GSM-8K}
& \textbf{SVAMP}
& \textbf{MM}
& \textbf{MATH}
& \textbf{AIME-24}
& \textbf{AIME-25}
& \textbf{MATH-500}
& \textbf{PolyMATH} \\
\midrule

\multicolumn{10}{l}{
\textbf{Tool-Augmented Baseline Framework Comparisons}
} \\
\midrule

\multirow{5}{*}{Qwen3-1.7B}
& CoT-FSP
& 69.00 & 67.71 & 41.03 & 57.41
& 20.00 & 16.67 & 46.29 & 23.81 \\
& CoT-FT
& 72.77 & 70.48 & 42.97 & 62.90
& 26.67 & 20.00 & 51.98 & 26.39 \\
& Toolformer
& 77.13 & 74.45 & 45.44 & 65.79
& 30.00 & 26.67 & 58.12 & 29.77 \\
& CoA
& \underline{79.54} & \underline{77.03} & \underline{48.13} & \underline{71.04}
& \underline{36.67} & \underline{36.67} & \underline{63.81} & \underline{33.12} \\

& ARCHER
& \textbf{83.72} & \textbf{81.53} & \textbf{51.58} & \textbf{78.06}
& \textbf{46.67} & \textbf{43.33} & \textbf{71.36} & \textbf{37.93} \\

\midrule

\multirow{5}{*}{Gemma4-E2B}
& CoT-FSP
& 71.74 & 70.45 & 44.07 & 62.82
& 50.00 & 43.33 & 51.39 & 26.41 \\
& CoT-FT
& 75.39 & 73.90 & 46.38 & 66.09
& 56.67 & 46.67 & 57.41 & 29.07 \\
& Toolformer
& 77.76 & 79.20 & 49.14 & 71.36
& \underline{60.00} & 46.67 & 64.95 & 32.63 \\
& CoA
& \underline{83.56} & \underline{83.16} & \underline{51.36} & \underline{78.42}
& \underline{60.00} & \underline{53.33} & \underline{70.50} & \underline{37.67} \\

& ARCHER
& \textbf{86.40} & \textbf{88.02} & \textbf{56.23} & \textbf{82.77}
& \textbf{66.67} & \textbf{60.00} & \textbf{80.50} & \textbf{41.45} \\

\midrule

\multirow{5}{*}{Qwen3.5-4B}
& CoT-FSP
& 77.63 & 75.64 & 46.81 & 67.88
& 36.67 & 30.00 & 58.02 & 29.03 \\
& CoT-FT
& 79.24 & 76.78 & 49.32 & 72.63
& 36.67 & 30.00 & 65.48 & 32.88 \\
& Toolformer
& 83.91 & 81.94 & 51.17 & 76.20
& 50.00 & 46.67 & 71.26 & 36.92 \\
& CoA
& \underline{85.85} & \underline{85.67} & \underline{55.22} & \underline{83.33}
& \underline{56.67} & \underline{53.33} & \underline{78.57} & \underline{41.35} \\

& ARCHER
& \textbf{87.89} & \textbf{89.87} & \textbf{60.21} & \textbf{90.82}
& \textbf{63.33} & \textbf{56.67} & \textbf{86.69} & \textbf{46.79} \\

\midrule

\multirow{5}{*}{\shortstack[l]{Falcon-H1R-7B}}
& CoT-FSP
& 73.61 & 77.99 & 52.56 & 70.53
& \underline{73.33} & 66.67 & 60.72 & 35.88 \\
& CoT-FT
& 75.16 & 83.52 & 56.51 & 74.74
& \underline{73.33} & 66.67 & 67.70 & 41.43 \\
& Toolformer
& 78.33 & 87.11 & 59.24 & 81.53
& \underline{73.33} & 66.67 & 73.31 & 45.50 \\
& CoA
& \underline{81.50} & \underline{92.67} & \underline{63.43} & \underline{86.89}
& \textbf{80.00} & \underline{73.33} & \underline{83.44} & \underline{52.02} \\

& ARCHER
& \textbf{85.96} & \textbf{96.29} & \textbf{66.64} & \textbf{92.74}
& \textbf{80.00} & \textbf{76.67} & \textbf{90.23} & \textbf{56.94} \\

\midrule

\multirow{5}{*}{\shortstack[l]{Qwen3-8B-thinking}}
& CoT-FSP
& 80.47 & 81.65 & 43.19 & 71.07
& 70.00 & 56.67 & 64.92 & 33.15 \\
& CoT-FT
& 83.17 & 85.33 & 45.98 & 76.54
& 70.00 & 56.67 & 70.19 & 36.81 \\
& Toolformer
& 84.86 & 87.89 & 50.11 & 84.42
& 70.00 & 56.67 & 80.21 & 42.73 \\
& CoA
& \underline{88.08} & \underline{92.42} & \underline{51.81} & \underline{88.34}
& \underline{73.33} & \underline{70.00} & \underline{86.32} & \underline{46.59} \\

& ARCHER
& \textbf{94.72} & \textbf{97.21} & \textbf{56.78} & \textbf{97.41}
& \textbf{76.67} & \textbf{76.67} & \textbf{98.60} & \textbf{52.48} \\

\midrule
\multicolumn{10}{l}{
\textbf{RL-specialized Reasoning Backbones}
} \\
\midrule

\multirow{3}{*}{\shortstack[l]{DeepSeek-R1-Distill-Qwen-7B}}
& Vanilla
& \underline{91.35} & 88.21 & 69.33 & 93.05 & 63.33 & 56.67 & 92.80 & 28.91 \\
& CoA
& 89.94 & \underline{90.15} & \underline{70.25} & \underline{94.01} & \underline{70.00} & \underline{60.00} & \underline{93.41} & \underline{30.15} \\

& ARCHER
& \textbf{94.79} & \textbf{92.51} & \textbf{73.38} & \textbf{96.25} & \textbf{80.00} & \textbf{66.67} & \textbf{93.66} & \textbf{37.04} \\

\midrule

\multirow{3}{*}{MiMo-7B-RL}
& Vanilla
& \underline{89.38} & 90.19 & 70.38 & 94.21 & 70.00 & \underline{63.33} & 94.60 & 31.25 \\
& CoA
& 88.21 & \underline{90.22} & \underline{71.85} & \underline{95.04} & \underline{76.67} & \underline{63.33} & \underline{95.02} & \underline{34.50} \\

& ARCHER
& \textbf{95.56} & \textbf{94.47} & \textbf{74.41} & \textbf{95.13} & \textbf{83.33} & \textbf{70.00} & \textbf{95.43} & \textbf{39.61} \\

\bottomrule
\end{tabular}%
}
\caption{Comparison of ARCHER with tool-augmented baseline frameworks on five standard backbones and with Vanilla and CoA variants on two reasoning-specialized backbones across eight mathematical reasoning benchmarks. \textbf{Boldface} indicate best values and \underline{underlines} indicate second-best values. All values are percentages.}

\label{tab:expanded_baseline_comparison}
\end{table*}

\paragraph{Baselines and evaluation protocol.}
For the five standard backbones, we compare ARCHER with CoT-FSP (Chain-of-Thought 3-shot Prompting without fine-tuning), CoT-FT (Chain-of-Thought Fine-Tuning), Toolformer \citep{schick2023toolformer}, and CoA \cite{gao2024efficient}; for the reasoning-specialized backbones, we compare Vanilla, CoA, and ARCHER. All results are obtained from direct runs on identical benchmark instances with a shared normalization and scoring pipeline, while each method retains its native reasoning or tool-interaction format. The two comparison blocks are reported separately because their baselines and training histories differ. Each model produces one response or trajectory per example. ARCHER follows the executable loop in Section~\ref{sec:methodology}, terminating at \texttt{ANSWER} or the step limit; missing or invalid terminal answers are incorrect. Other methods use benchmark-specific answer extraction. Numeric, symbolic, and multiple-choice outputs are normalized before scoring.

\paragraph{Metrics.}
Final-answer accuracy is the primary metric, with AIME results reported as single-run exact match over the complete sets. On GSM-8K, we additionally report Tool Score, tool precision and recall, observation consistency, Process Score, valid trace rate, trace edit distance, and reward-family diagnostics. Tool Score and Process Score are dataset-level macro-averages; valid trace rate measures conformity to the action schema, and observation consistency measures agreement with deterministic execution.

\section{Results}
\label{sec:results}

We evaluate ARCHER along two dimensions: whether executable supervision improves reasoning correctness and trajectory validity on GSM-8K, and whether these gains transfer across model families and mathematical benchmarks.

\paragraph{ARCHER generalizes across model families.}
In the five-backbone comparison in Table~\ref{tab:expanded_baseline_comparison}, ARCHER is strictly best in 39 of 40 backbone-benchmark combinations and tied for best on Falcon-H1R-7B for AIME-2024. Its mean score across the 40 cells is 73.36, compared with 67.85 for CoA, yielding an average improvement of 5.51 percentage points. Averaged across eight benchmarks, the gains over CoA are 6.02 points for Qwen3-1.7B, 5.50 for Gemma4-E2B, 5.29 for Qwen3.5-4B, 4.02 for Falcon-H1R-7B, and 6.71 for Qwen3-8B-thinking. The improvement therefore remains consistent across model families, sizes, and reasoning-oriented initializations.

\paragraph{Executable grounding complements reasoning-specialized models.}
ARCHER also outperforms both Vanilla and CoA on every benchmark for DeepSeek-R1-Distill-Qwen-7B and MiMo-7B-RL. Relative to Vanilla, AIME-2024 accuracy increases from 63.33 to 80.00 for DeepSeek and from 70.00 to 83.33 for MiMo. PolyMATH improves from 28.91 to 37.04 and from 31.25 to 39.61, respectively, while gains on MATH-500 are smaller but remain positive. These results indicate that deterministic execution is not merely compensation for weak arithmetic ability. It provides additional reliability even when the backbone already exhibits strong reasoning behavior from reinforcement learning or distillation.

\paragraph{Improvements are largest on formal, multi-step benchmarks.}
On the natural-language word-problem benchmarks GSM-8K and SVAMP, ARCHER averages 87.74 and 90.58 across the five standard backbones, improving over CoA by 4.03 and 4.39 points. Larger margins appear on benchmarks requiring longer symbolic derivations. ARCHER improves over CoA by 6.76 points on MATH and 8.95 points on MATH-500, with gains of 5.33 and 5.34 points on AIME-2024 and AIME-2025. This pattern suggests that action structure, runtime observations, and trajectory refinement support not only arithmetic execution but also strategy selection and consistency across extended solutions.

\paragraph{Language and knowledge remain distinct challenges.}
ARCHER averages 58.29 on MM, improving over CoA by 4.30 points, but absolute performance remains below that on the arithmetic and competition benchmarks. This reflects the broader knowledge, interpretation, and multiple-choice reasoning required by MM. PolyMATH produces the lowest ARCHER average, 47.12, while still improving over CoA by 4.97 points. The result demonstrates multilingual transfer, but also shows that calculator grounding does not remove difficulties caused by language variation. More detailed language-wise analysis is needed to determine whether these gains are uniformly distributed.

\begin{figure}[t]
    \centering
    \includegraphics[width=0.95\columnwidth]{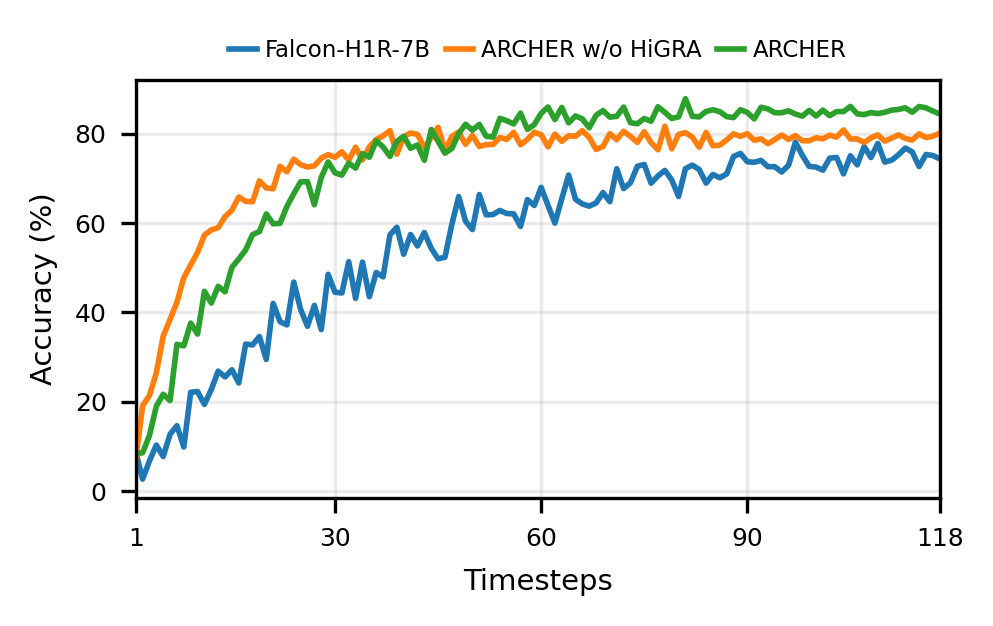}
    \caption{Training dynamics study of ARCHER.}
    \label{fig:tri_compare}
\end{figure}

\paragraph{Backbone specialization and training stages remain important.}
Average ARCHER performance increases from 61.77 on Qwen3-1.7B to 70.26 on Gemma4-E2B, 72.78 on Qwen3.5-4B, 80.68 on Falcon-H1R-7B, and 81.32 on Qwen3-8B-thinking. However, benchmark-level rankings do not follow parameter count alone. Qwen3-8B-thinking performs best on GSM-8K, SVAMP, MATH, and MATH-500, whereas Falcon-H1R-7B is strongest on MM, AIME-2024, and PolyMATH, with both models tied on AIME-2025. Figure~\ref{fig:tri_compare} further shows that TRACE provides rapid early improvement, while HiGRA continues improving after the supervised model begins to plateau. Together, these results support the two-stage design: TRACE establishes structured executable reasoning, and HiGRA converts that structure into stronger accuracy, grounding, and trajectory validity.

\section{Ablation Study}
\label{sec:ablations_error_analysis}

\begin{table}[t]
\centering
\scriptsize
\setlength{\tabcolsep}{2.5pt}

\begin{tabular*}{\columnwidth}{
@{\extracolsep{\fill}}lccc@{}
}
\toprule
\textbf{Metric}
& \textbf{Falcon-H1R-7B}
& \textbf{ARCHER w/o HiGRA}
& \textbf{ARCHER} \\
\midrule
\textbf{Acc.$\uparrow$}
& 77.40 & 79.06 & \textbf{85.96} \\
\textbf{\shortstack[l]{Tool Score$\uparrow$}}
& 0.00 & 0.46 & \textbf{0.76} \\
\textbf{\shortstack[l]{Tool Recall$\uparrow$}}
& 0.00 & 0.58 & \textbf{0.79} \\
\textbf{\shortstack[l]{Proc. Score$\uparrow$}}
& 0.52 & 0.67 & \textbf{0.79} \\
\textbf{\shortstack[l]{Valid Trace$\uparrow$}}
& 41.2 & 68.7 & \textbf{82.4} \\
\textbf{\shortstack[l]{Obs. Cons.$\uparrow$}}
& 0.00 & 0.84 & \textbf{0.93} \\
\bottomrule
\end{tabular*}

\caption{Stage-wise executable diagnostics on GSM-8K with Falcon-H1R-7B backbone. Accuracy and Valid Trace are percentages; all other metrics are reported on $[0,1]$. Obs. Cons. denotes observation consistency.}
\label{tab:gsm8k_stagewise_diagnostics}
\end{table}

\begin{table}[t]
\centering
\scriptsize
\setlength{\tabcolsep}{2.5pt}

\begin{tabularx}{\columnwidth}{@{}Lccc@{}}
\toprule
\textbf{Model Variant}
& \textbf{Acc.$\uparrow$}
& \textbf{\shortstack[c]{Tool Score$\uparrow$}}
& \textbf{\shortstack[c]{Proc. Score$\uparrow$}} \\
\midrule
Falcon-H1R-7B
& 77.40 & 0.00 & 0.52 \\
CoT-SFT
& 78.12 & 0.03 & 0.55 \\
ARCHER w/o HiGRA
& 79.06 & 0.46 & 0.67 \\
ARCHER w/ outcome-only GRPO
& 82.35 & 0.55 & 0.69 \\
ARCHER w/ flat micro-reward GRPO
& 83.10 & 0.64 & 0.72 \\
\midrule
ARCHER w/o tool reward
& 81.74 & 0.51 & 0.76 \\
ARCHER w/o process reward
& 82.18 & 0.71 & 0.61 \\
ARCHER w/o alignment reward
& 84.12 & 0.74 & 0.77 \\
ARCHER w/o gating
& 82.95 & 0.69 & 0.72 \\

\textbf{ARCHER}
& \textbf{85.96} & \textbf{0.76} & \textbf{0.79} \\
\bottomrule
\end{tabularx}

\caption{ Training and reward ablations on GSM-8K with Falcon-H1R-7B backbone. Training-target and HiGRA reward ablations. Accuracy is a percentage; all other metrics are reported on $[0,1]$.}
\label{tab:gsm8k_training_reward_ablations}
\end{table}

All ablations use Falcon-H1R-7B on GSM-8K, whose annotated arithmetic solutions enable detailed evaluation of tool use and executable trajectory quality. We isolate the effects of TRACE, HiGRA, alternative reward formulations, and individual reward components in Table~\ref{tab:gsm8k_training_reward_ablations}. Additional executable-trajectory diagnostics are provided in Appendix~\ref{app:additional_results}.

\paragraph{TRACE establishes executable behavior, while HiGRA drives accuracy.}
Table~\ref{tab:gsm8k_stagewise_diagnostics} separates the contributions of the two stages. TRACE increases accuracy from 77.40 to 79.06 while producing larger improvements in trajectory quality: valid trace rate rises from 41.2 to 68.7, Tool Score from 0.00 to 0.46, and Process Score from 0.52 to 0.67; tool recall and observation consistency reach 0.58 and 0.84. Full ARCHER increases accuracy to 85.96, raises valid trace rate to 82.4, and improves observation consistency to 0.93. Thus, TRACE establishes the executable interface, while HiGRA improves its correct and consistent use.

\paragraph{Training target and reward design.}
Table~\ref{tab:gsm8k_training_reward_ablations} shows that CoT-SFT provides only a small accuracy gain over the base model, from 77.40 to 78.12, with little improvement in tool or process scores. ARCHER w/o HiGRA reaches 79.06 and produces substantially stronger executable behavior, showing that the benefit of Stage~1 comes from replacing free-form rationale imitation with executable action supervision. Reward-guided refinement further improves performance: outcome-only GRPO reaches 82.35 and flat micro-reward GRPO reaches 83.10, whereas full ARCHER achieves 85.96. This supports HiGRA's grouped rewards and conditional gating over outcome-only or flat reward formulations.

\paragraph{Component contributions and gating.}
Removing the tool reward causes the largest component-level accuracy drop, from 85.96 to 81.74, confirming that explicit tool grounding is central to ARCHER's gains. Removing the process reward lowers accuracy to 82.18 and Process Score from 0.79 to 0.61, while removing alignment has the smallest effect, reducing accuracy to 84.12 and supporting its role as a weak stabilizer. Removing gating lowers accuracy to 82.95 and reduces both tool and process scores, indicating that auxiliary reward families should contribute only when relevant; otherwise, they can introduce noisy or conflicting optimization pressure.

\section{Conclusion and Future Work}

We presented \textbf{ARCHER}, a two-stage framework combining executable trace supervision with reward-guided refinement. Stage~1 compiles free-form rationales into typed action traces containing calculator calls, runtime observations, verification actions, and final answers, while Stage~2 refines this policy with grouped and conditionally gated rewards over correctness, tool grounding, process discipline, and trace alignment. On Falcon-H1R-7B, ARCHER improves GSM-8K accuracy from 77.40 to 85.96 while also improving tool recall, observation consistency, valid trace rate, and process quality. Across five standard backbones, ARCHER is best or tied-best in all 40 model-benchmark comparisons and improves over the strongest tool-augmented baseline by 5.51 points on average. It also improves both DeepSeek-R1-Distill-Qwen-7B and MiMo-7B-RL across all eight benchmarks, showing that deterministic calculator execution remains complementary to reasoning capabilities induced through reinforcement learning or distillation.

Future work can extend ARCHER in three directions. First, the action space can be expanded beyond calculator calls to include symbolic solvers, unit converters, code execution, retrieval, and proof checking, enabling richer forms of executable reasoning. Second, TRACE can be generalized from rule-based arithmetic trace compilation to broader trace construction pipelines that handle more diverse rationale formats and reasoning domains. Third, HiGRA can be further improved through automatic reward calibration, adaptive gating, and model-specific reward weighting. These extensions would test whether the same principle behind ARCHER, aligning reasoning supervision with executable intermediate computation, can improve SLMs across a wider range of tasks.

\section*{Limitations}

ARCHER is evaluated primarily in mathematical reasoning settings where intermediate computations can be executed and final answers can be checked reliably. This makes the setting well suited for studying calculator-grounded reasoning, but it also means that the current results should be interpreted most directly for tasks with verifiable numerical or symbolic structure. The current implementation uses a deterministic TRACE compiler and manually designed HiGRA reward families, so broader domains may require richer action schemas, additional tools, and task-specific reward calibration. We view these as natural extensions of the framework rather than fundamental restrictions, since ARCHER is designed around the general idea of training compact models to produce executable and auditable reasoning trajectories.

\section*{Disclosure of AI Assistant Use}
The authors used LLM-based writing assistants to revise and improve portions of the manuscript prose based on text originally written by the authors. The authors retained full responsibility for the scientific content of the paper, including the research ideas, technical approach, experiments, results, and conclusions. All AI-assisted revisions were carefully inspected by the authors for accuracy, and the manuscript was manually checked to ensure that no hallucinated, fabricated, or unsupported content was introduced. GPT-Codex and Claude Code Coding Agents were used to assist with implementation and debugging for some parts of the experimental codebase.

\section*{Ethical Considerations}
ARCHER is a research framework for mathematical reasoning and is not intended for safety-critical decision making. Although executable traces improve inspectability, they may still contain incorrect decompositions, calculator inputs, or uses of correct observations. Apparent trace structure should therefore not be treated as a guarantee of correctness, and outputs should be independently verified in high-stakes settings. Training and evaluating multiple language models also incurs computational and environmental costs, although all experiments in this study were conducted locally on a single Apple M5 Pro system without external compute clusters. The study uses existing mathematical reasoning benchmarks and does not collect personal data.

\bibliography{custom}

\clearpage
\appendix

\numberwithin{equation}{section}
\numberwithin{figure}{section}
\numberwithin{table}{section}

\section{Full ARCHER Methodology}
\label{app:archer_details}

This appendix gives the full technical description of ARCHER: \textbf{A}ction-trace \textbf{R}einforcement for \textbf{C}alculator-grounded \textbf{H}ierarchical \textbf{E}xecutable \textbf{R}easoning. ARCHER consists of two stages. TRACE constructs executable supervision from legacy rationales and trains the initial policy. HiGRA refines the TRACE policy with grouped and gated trajectory rewards.

\subsection{Problem Setting}

Each training example is represented as $(x,r,y^\star)$, where $x$ is the input problem, $r$ is a free-form rationale or chain-of-thought solution, and $y^\star$ is the gold final answer. The goal is to train a compact student policy $\pi_\theta$ that generates a structured trajectory $\tau$ and final answer $\hat{y}$:
\[
\pi_\theta: x \mapsto (\tau,\hat{y}).
\]
Unlike ordinary rationale distillation, ARCHER does not train the student to imitate raw explanatory text. Instead, it trains the student to operate in an executable action language with explicit calculator calls and deterministic runtime observations.

\subsection{Executable Action Language}

A trajectory is a sequence of structured states:
\[
\tau=(s_1,s_2,\dots,s_T),
\qquad
s_t=(a_t,u_t,o_t).
\]
Here, $a_t$ is an action type, $u_t$ is the model-emitted payload, and $o_t$ is the observation returned by the runtime when the action invokes a tool. ARCHER uses the action vocabulary
\[
\mathcal{A} =
\left\{
\begin{aligned}
&\texttt{READ},\ \texttt{PLAN},\ \texttt{CALL\_CALC},\\
&\texttt{OBSERVE},\ \texttt{VERIFY},\ \texttt{ANSWER}
\end{aligned}
\right\}
\]

\begin{table}[t]
\centering
\small

\setlength{\tabcolsep}{3.5pt}

\begin{tabularx}{\columnwidth}{@{}l>{\raggedright\arraybackslash}X@{}}
\toprule
\textbf{Action} & \textbf{Role} \\
\midrule
\texttt{READ} & Restates or identifies the relevant information in the problem. \\
\texttt{PLAN} & Specifies the next reasoning step or decomposition. \\
\texttt{CALL\_CALC} & Submits an arithmetic expression to the calculator runtime. \\
\texttt{OBSERVE} & Records the deterministic output returned by the runtime. \\
\texttt{VERIFY} & Checks whether the intermediate or final result is consistent. \\
\texttt{ANSWER} & Emits the final normalized answer. \\
\bottomrule
\end{tabularx}

\caption{ARCHER action vocabulary. The model generates reasoning actions and calculator-call payloads; calculator observations are produced by the runtime.}
\label{tab:archer_action_vocab}
\end{table}

The final prediction is extracted from the terminal answer action:
\[
\hat{y}=\mathrm{Ans}(\tau).
\]

\subsection{Stage 1: TRACE}

TRACE is the supervised compilation stage of ARCHER. It converts a free-form rationale into a calculator-grounded executable trace:
\[
\tau^\star=\mathcal{C}(x,r,y^\star),
\]
where $\mathcal{C}$ is the TRACE compiler.

\subsubsection{Trace Compilation}

The compiler segments the rationale into textual reasoning spans and arithmetic spans. Textual spans are mapped to reasoning actions such as \texttt{READ}, \texttt{PLAN}, and \texttt{VERIFY}. Arithmetic spans are normalized into executable calculator expressions. For each extracted expression $e_t$, TRACE emits:
\[
\begin{aligned}
s_t^\star &= (\texttt{CALL\_CALC}, e_t, \varnothing), \\
s_{t+1}^\star &= (\texttt{OBSERVE}, \varnothing, f_{\mathrm{calc}}(e_t)).
\end{aligned}
\]
Here, $f_{\mathrm{calc}}$ is the deterministic calculator runtime. For example, a rationale fragment containing the operation $48/6=8$ is converted into:
\[
\texttt{CALL\_CALC}(48/6)
\rightarrow
\texttt{OBSERVE}(8).
\]

For GSM-8K, the compiler uses the dataset's inline calculator format, \texttt{<< expr = result >>}, to identify arithmetic spans. Each expression is normalized, executed, and paired with the resulting observation. Text outside calculator annotations is retained as reasoning content and mapped into structured actions.

\subsubsection{Compiler Policy}

The TRACE compiler is deterministic and rule-based. It does not use an external LLM to rewrite or repair rationales. The compiler follows a retain-and-annotate policy:

\begin{itemize}
    \item Examples with valid calculator spans are converted into executable traces.
    \item Examples without calculator spans are still converted into structured traces when possible.
    \item Arithmetic mismatches are repaired at the observation level when exact execution succeeds.
    \item Unsupported calculator spans are retained with compiler metadata.
    \item Each trace is terminated with the normalized gold answer.
\end{itemize}

This policy avoids discarding useful training examples while preserving metadata for later filtering and analysis.

\subsubsection{Calculator Runtime}

The calculator runtime separates numerical execution from reasoning control. If the model emits a calculator call with payload $e_t$, the runtime returns:
\[
o_t=f_{\mathrm{calc}}(e_t).
\]
The observation is inserted into the transcript and becomes part of the context for subsequent model actions. The model therefore learns what to compute and how to use the result, while the runtime performs exact arithmetic.

\subsubsection{Student Policy}

The student policy generates action types and payloads autoregressively:
\[
\pi_\theta(a_t,u_t \mid x,s_{<t}).
\]
The trajectory distribution is:
\[
\pi_\theta(\tau \mid x)
=
\prod_{t \in \mathcal{M}(\tau)}
\pi_\theta(a_t,u_t \mid x,s_{<t}),
\]
where $\mathcal{M}(\tau)$ is the set of model-generated positions. Runtime observations are not sampled from the language model and are not treated as model decisions.

\subsubsection{TRACE Supervised Fine-tuning}

Given a compiled gold trace $\tau^\star=(s_1^\star,\dots,s_{T^\star}^\star)$, TRACE trains the student with:
\[
\begin{aligned}
\mathcal{L}_{\mathrm{TRACE}}(\theta)
&= -\sum_{t \in \mathcal{M}(\tau^\star)} \\
&\quad \log \pi_\theta(a_t^\star,u_t^\star \mid x,s_{<t}^\star).
\end{aligned}
\]
Runtime observations are included in the teacher-forced context, but the loss is applied only to model-generated positions. This preserves the distinction between decisions made by the policy and observations produced by external execution.

\subsubsection{TRACE Inference Loop}

At inference time, the TRACE-trained policy runs iteratively. At step $t$, the model emits an action and payload:
\[
(a_t,u_t)\sim \pi_\theta(\cdot \mid x,s_{<t}).
\]
The runtime observation is:
\[
o_t =
\begin{cases}
f_{\mathrm{calc}}(u_t), & \text{if } a_t=\texttt{CALL\_CALC}, \\
\varnothing, & \text{otherwise}.
\end{cases}
\]
The resulting state is appended to the trajectory. The process continues until the model emits \texttt{ANSWER} or reaches the maximum step budget.

\subsection{Stage 2: HiGRA}

HiGRA is the reward-guided refinement stage of ARCHER. It starts from the TRACE policy and improves it using trajectory-level rewards. The purpose of HiGRA is to correct failures that supervised trace imitation alone may not remove, including missing calculator calls, wrong expressions, ignored observations, malformed traces, and premature answers.

\subsubsection{Reward Families}

HiGRA groups fine-grained feedback into four macro-reward families:
\[
R_{\mathrm{out}},
\qquad
R_{\mathrm{tool}},
\qquad
R_{\mathrm{proc}},
\qquad
R_{\mathrm{align}}.
\]
The outcome reward measures final-answer correctness. The tool reward measures calculator use and execution faithfulness. The process reward measures trajectory validity and discipline. The alignment reward softly encourages compatibility with the compiled TRACE supervision.

\subsubsection{Outcome Reward}

The outcome reward is:
\[
R_{\mathrm{out}}(\tau)
=
\mathbf{1}[\hat{y}=y^\star].
\]
This is the primary task reward and anchors optimization to answer correctness.

\subsubsection{Tool-grounding Reward and Tool Score}

Let $c^\star$ be the number of calculator calls in the compiled
reference trace $\tau^\star$, and let $m$ be the number of
\texttt{CALL\_CALC} actions emitted by the model in trajectory
$\tau$. Define
\[
u_{\mathrm{need}}=\mathbf{1}[c^\star>0],
\qquad
u_{\mathrm{used}}=\mathbf{1}[m>0].
\]

For the emitted calls, let $e_i$ be the payload of call $i$ and let
$o_i$ be its parser-paired observation, when present. Define
\[
\begin{aligned}
u_{\mathrm{obs}}
&= \frac{1}{\max(m,1)}
\sum_{i=1}^{m} \\
&\quad \mathbf{1}[\text{call } i \text{ has a paired observation}].
\end{aligned}
\]
and
\[
u_{\mathrm{cons}}
=
\frac{1}{\max(m,1)}
\sum_{i=1}^{m}
\mathbf{1}[o_i=f_{\mathrm{calc}}(e_i)].
\]

The per-trajectory tool reward is
\begin{equation}
\begin{aligned}
R_{\mathrm{tool}}(\tau)
&=0.2u_{\mathrm{used}}+0.2u_{\mathrm{obs}}\\
&\quad+0.6u_{\mathrm{cons}},
&& u_{\mathrm{need}}=1,\\[2pt]
R_{\mathrm{tool}}(\tau)
&=1-0.2u_{\mathrm{used}},
&& u_{\mathrm{need}}=0.
\end{aligned}
\label{eq:app_tool_reward_operational}
\end{equation}

Thus, when the compiled reference trace requires calculator use,
the reward assigns credit for emitting at least one calculator call,
pairing emitted calls with observations, and maintaining consistency
with deterministic calculator execution. When the reference trace
does not require calculator use, a trajectory receives full tool
credit for avoiding the calculator and a penalty of $0.2$ for
emitting one or more unnecessary calls.

Every emitted \texttt{CALL\_CALC} action contributes to $m$,
including malformed, empty, and non-executable calls. Such calls
receive no consistency credit. A parser-paired runtime-error
observation can receive observation-presence credit, but it receives
no consistency credit. The consistency indicator is defined to be
zero whenever the submitted expression cannot be parsed or executed.

For an evaluation set $D$, the reported Tool Score is the
macro-average
\begin{equation}
\mathrm{ToolScore}(D)
=
\frac{1}{|D|}
\sum_{\tau\in D}R_{\mathrm{tool}}(\tau).
\label{eq:app_tool_score}
\end{equation}

\subsubsection{Process-discipline Reward and Process Score}

Process discipline separates structural-format quality from
verification presence, trajectory length, final-answer presence, and
premature termination. Let $\mathcal{K}(\tau)$ denote the set of
structural checks applicable to trajectory $\tau$, let
$K(\tau)=|\mathcal{K}(\tau)|$, and let
\[
q_k(\tau)\in\{0,1\}
\]
indicate whether applicable structural check $k$ passes. The
structural-format score is
\begin{equation}
p_{\mathrm{fmt}}(\tau)
=
\frac{1}{K(\tau)}
\sum_{k\in\mathcal{K}(\tau)}q_k(\tau).
\label{eq:app_format_score}
\end{equation}

The structural checks cover:

\begin{itemize}
    \item recognized and parseable action tags;
    \item valid payloads for the corresponding action types;
    \item nonempty and parseable \texttt{CALL\_CALC} expressions;
    \item exactly one paired \texttt{OBSERVE} action for each calculator call;
    \item no orphan or duplicated observations;
    \item at most one \texttt{ANSWER} action;
    \item a terminal \texttt{ANSWER} action when an answer is present;
    \item no unknown action tags; and
    \item no unparsed residual text.
\end{itemize}

The implementation applies at least one structural check to every
trajectory, so $K(\tau)\geq 1$. The format score measures structural
format only. Verification presence, answer presence, length, and
premature termination are represented separately:
\[
\begin{aligned}
p_{\mathrm{verify}}(\tau)
&=
\mathbf{1}[\texttt{VERIFY}\in\tau],\\
p_{\mathrm{short}}(\tau)
&=
\mathbf{1}[T\leq T_{\max}],\\
p_{\mathrm{final}}(\tau)
&=
\mathbf{1}[\texttt{ANSWER}\in\tau],\\
p_{\mathrm{prem}}(\tau)
&=
\mathbf{1}[\text{premature final answer}].
\end{aligned}
\]

The per-trajectory process reward is
\begin{equation}
\begin{aligned}
q_{\mathrm{proc}}(\tau)
&=0.4p_{\mathrm{fmt}}(\tau)+0.2p_{\mathrm{verify}}(\tau)\\
&\quad+0.2p_{\mathrm{short}}(\tau)+0.2p_{\mathrm{final}}(\tau)\\
&\quad-0.5p_{\mathrm{prem}}(\tau),\\
R_{\mathrm{proc}}(\tau)
&=\operatorname{clip}_{[0,1]}\!\left(q_{\mathrm{proc}}(\tau)\right).
\end{aligned}
\label{eq:app_process_reward_operational}
\end{equation}
where
\[
\operatorname{clip}_{[0,1]}(z)
=
\min(1,\max(0,z)).
\]

The binary shortness component is part of the process-discipline
reward. The separate soft length penalty $P_{\mathrm{len}}$ in the
total HiGRA reward measures how far an overlong trajectory exceeds
the preferred trajectory budget.

For an evaluation set $D$, the reported Process Score is the
macro-average
\begin{equation}
\mathrm{ProcessScore}(D)
=
\frac{1}{|D|}
\sum_{\tau\in D}R_{\mathrm{proc}}(\tau).
\label{eq:app_process_score}
\end{equation}

\subsubsection{Trace-alignment Reward}

Trace alignment softly regularizes the rollout toward the compiled executable trace without requiring exact imitation. Let $\mathcal{S}(\tau)$ and $\mathcal{S}^\star$ denote the generated and gold action schemas, and let $\mathcal{O}(\tau)$ and $\mathcal{O}^\star$ denote the generated and gold operation sequences. Define:
\[
a_{\mathrm{schema}}
=
\frac{|\mathcal{S}(\tau)\cap\mathcal{S}^\star|}
{|\mathcal{S}^\star|+\varepsilon},
\]
\[
a_{\mathrm{op}}
=
\frac{|\mathcal{O}(\tau)\cap\mathcal{O}^\star|}
{|\mathcal{O}^\star|+\varepsilon},
\]
and
\[
\begin{aligned}
a_{\mathrm{verify}}
&=
\mathbf{1}[\text{gold trace expects verification}]
\\
&\quad{}\cdot
\mathbf{1}[\texttt{VERIFY}\in\tau].
\end{aligned}
\]
The alignment reward is:
\[
R_{\mathrm{align}}(\tau)
=
\delta_1a_{\mathrm{schema}}
+
\delta_2a_{\mathrm{op}}
+
\delta_3a_{\mathrm{verify}}.
\]
This term is intentionally weaker than outcome, tool, and process rewards.

\subsubsection{Hierarchical Gating}

HiGRA gates auxiliary rewards so that irrelevant reward dimensions do not contribute uniformly to every trajectory. Tool grounding is naturally gated by tool necessity through $u_{\mathrm{need}}$. Alignment is also gated:
\[
\widetilde{R}_{\mathrm{align}}(\tau)
=
g_{\mathrm{align}}(\tau)R_{\mathrm{align}}(\tau),
\]
where
\[
g_{\mathrm{align}}(\tau)
=
\mathbf{1}
[
R_{\mathrm{out}}(\tau)=1
\vee
R_{\mathrm{tool}}(\tau)>\eta
].
\]
Thus, alignment is emphasized only when the trajectory is already correct or shows sufficient tool-grounded structure.

\subsubsection{Length and Loop Penalty}

To discourage repetitive or overly long trajectories, HiGRA uses a soft length penalty:
\[
P_{\mathrm{len}}(\tau)
=
\lambda_{\mathrm{len}}
\max\left(0,\frac{T-T_{\max}}{T_{\max}}\right).
\]
This is smoother than a hard failure condition and is better suited to compact policies.

\subsubsection{Final HiGRA Reward}

The final HiGRA reward is:
\[
\begin{aligned}
R_{\mathrm{HiGRA}}(\tau)
&=w_{\mathrm{out}}R_{\mathrm{out}}(\tau)\\
&\quad+w_{\mathrm{tool}}R_{\mathrm{tool}}(\tau)\\
&\quad+w_{\mathrm{proc}}R_{\mathrm{proc}}(\tau)\\
&\quad+w_{\mathrm{align}}\widetilde{R}_{\mathrm{align}}(\tau)\\
&\quad-P_{\mathrm{len}}(\tau).
\end{aligned}
\]
The weights follow the priority ordering:
\[
w_{\mathrm{out}}
>
w_{\mathrm{tool}}
>
w_{\mathrm{proc}}
>
w_{\mathrm{align}}.
\]
In our experiments, we use:
\[
\begin{aligned}
w_{\mathrm{out}} &= 2.0,
&\qquad
w_{\mathrm{tool}} &= 1.0,\\
w_{\mathrm{proc}} &= 0.5,
&\qquad
w_{\mathrm{align}} &= 0.25.
\end{aligned}
\]

\subsection{Policy Optimization with GRPO}

HiGRA starts from the TRACE-supervised policy $\pi_{\theta}^{\mathrm{SFT}}$. Given an input $x$, the current policy samples a group of $G$ trajectories:
\[
\{\tau^{(1)},\dots,\tau^{(G)}\}
\sim
\pi_{\phi_{\mathrm{old}}}(\cdot\mid x).
\]
Each trajectory is executed through the calculator runtime and scored:
\[
r_i=R_{\mathrm{HiGRA}}(\tau^{(i)}).
\]
The group mean and standard deviation are:
\[
\begin{aligned}
\bar{r} &= \frac{1}{G}\sum_{i=1}^{G}r_i, \\
\sigma_r &= \sqrt{\frac{1}{G}\sum_{i=1}^{G}(r_i-\bar{r})^2+\epsilon}.
\end{aligned}
\]
The group-relative advantage is:
\[
A_i=\frac{r_i-\bar{r}}{\sigma_r}.
\]

For token $z_{i,t}$ in trajectory $\tau^{(i)}$, define the importance ratio:
\[
\rho_{i,t}
=
\frac{
\pi_{\phi}(z_{i,t}\mid x,z_{i,<t})
}{
\pi_{\phi_{\mathrm{old}}}(z_{i,t}\mid x,z_{i,<t})
}.
\]
The clipped ratio is:
\[
\bar{\rho}_{i,t}
=
\mathrm{clip}
(
\rho_{i,t},
1-\varepsilon,
1+\varepsilon
).
\]
The GRPO loss is:
\[
\begin{aligned}
\mathcal{L}_{\mathrm{GRPO}}(\phi)
&=
-\mathbb{E}_{x}
\Bigg[
\frac{1}{G}
\sum_{i=1}^{G}
\frac{1}{|\tau^{(i)}|}
\sum_{t=1}^{|\tau^{(i)}|} \\
&\qquad\qquad
\min
\left(
\rho_{i,t}A_i,
\bar{\rho}_{i,t}A_i
\right)
\Bigg].
\end{aligned}
\]
This objective increases the likelihood of trajectories whose HiGRA reward is higher than the group average while clipping overly large policy updates.

\section{Additional ARCHER Diagnostic Results}
\label{app:additional_results}

This appendix reports additional diagnostics for ARCHER that support the main results but are not required in the main body. The goal is to separate answer-level gains from improvements in executable reasoning behavior, including trace validity, calculator use, observation consistency, first-error position, and reward-family behavior. In these diagnostic tables, \textit{w/o HiGRA} denotes the Stage 1-only ARCHER variant.

\paragraph{Structural and tool-use metrics.}
Table~\ref{tab:trace_tool_metrics} reports whether the model produces executable traces and uses calculator calls faithfully. Accuracy measures final-answer correctness. NED$_{\mathrm{act}}$ and NED$_{\mathrm{op}}$ are normalized edit distances to the compiled gold action and operation sequences, so lower values indicate closer alignment with the target executable trace. Valid Trace is the percentage of outputs that follow the expected action format. Tool precision measures whether emitted tool calls are appropriate, tool recall measures whether needed calls are made, and observation consistency measures agreement between recorded observations and deterministic calculator outputs.

\begin{table*}[t]
\centering
\small

\setlength{\tabcolsep}{1mm}
\begin{tabular}{lccc}
\toprule
\textbf{Metric} & \textbf{Falcon} & \textbf{\shortstack[c]{ARCHER\\w/o HiGRA}} & \textbf{ARCHER} \\
\midrule
Acc.$\uparrow$ & 77.40 & 79.06 & \textbf{85.96} \\
NED$_{\mathrm{act}}\downarrow$ & 0.48 & 0.29 & \textbf{0.21} \\
NED$_{\mathrm{op}}\downarrow$ & 0.55 & 0.34 & \textbf{0.26} \\
Valid Trace$\uparrow$ & 41.2 & 68.7 & \textbf{82.4} \\
Tool Prec.$\uparrow$ & 0.00 & 0.71 & \textbf{0.83} \\
Tool Rec.$\uparrow$ & 0.00 & 0.58 & \textbf{0.79} \\
Obs. Cons.$\uparrow$ & 0.00 & 0.84 & \textbf{0.93} \\
\bottomrule
\end{tabular}
\caption{Structural and tool-use diagnostics on GSM-8K. TRACE substantially improves trace validity and tool use over the base model, while HiGRA further improves both executable structure and tool faithfulness.}
\label{tab:trace_tool_metrics}
\end{table*}

The main inference is that TRACE changes the form of the model's reasoning: valid traces increase from 41.2 to 68.7, and tool recall rises from 0.00 to 0.58. HiGRA then improves reliability within that format, raising valid trace rate to 82.4, tool recall to 0.79, and observation consistency to 0.93. Thus, the final accuracy gain is accompanied by cleaner execution behavior.

\paragraph{First-error position.}
Table~\ref{tab:first_error_position} shows where trajectories first become incorrect. No Error is the percentage of examples without a detected reasoning failure. Early Error refers to failures during problem understanding or planning. Tool-Step Error refers to failures in calculator-call construction or observation handling. Post-Tool Error refers to failures after a correct tool observation has already been produced. Mean First-Error Pos. is normalized from 0 to 1, where larger values mean the first error occurs later in the trajectory.

\begin{table*}[t]
\centering
\small

\setlength{\tabcolsep}{1mm}
\begin{tabular}{lccc}
\toprule
\textbf{Metric} 
& \textbf{Falcon} 
& \textbf{\shortstack[c]{ARCHER\\w/o HiGRA}}
& \textbf{ARCHER} \\
\midrule
No Error$\uparrow$ & 77.40 & 79.06 & \textbf{85.96} \\
Early Error$\downarrow$ & 12.3 & 8.64 & \textbf{4.84} \\
Tool-Step Error$\downarrow$ & 4.9 & 7.1 & \textbf{4.3} \\
Post-Tool Error$\downarrow$ & 5.4 & 5.2 & \textbf{4.9} \\
Mean First-Error Pos.$\uparrow$ & 0.38 & 0.51 & \textbf{0.64} \\
\bottomrule
\end{tabular}
\caption{First-error position on GSM-8K. TRACE shifts errors later in the trajectory, and HiGRA further reduces early and tool-step failures.}
\label{tab:first_error_position}
\end{table*}

The key pattern is that TRACE reduces early failures and moves the first error later, but it also exposes more tool-step errors because the model now attempts explicit calculator use. HiGRA reduces both early errors and tool-step errors, and the mean first-error position increases from 0.38 to 0.64. This indicates that ARCHER improves not only final correctness but also the durability of the reasoning process.

\paragraph{Reward-family correlation.}
Table~\ref{tab:reward_correlation} measures how strongly each reward family is associated with final-answer correctness. Correlation measures linear association with correctness, while AUC measures how well a reward score separates correct from incorrect trajectories. $R_{\mathrm{out}}$ has perfect association because it directly encodes answer correctness. The auxiliary rewards are not expected to be perfect predictors, but they should still correlate with better trajectories.

\begin{table}[t]
\centering
\small

\setlength{\tabcolsep}{5pt}
\begin{tabular}{lcc}
\toprule
\textbf{Reward} & \textbf{Corr.$\uparrow$} & \textbf{AUC$\uparrow$} \\
\midrule
$R_{\mathrm{out}}$ & 1.00 & 1.00 \\
$R_{\mathrm{tool}}$ & 0.54 & 0.78 \\
$R_{\mathrm{proc}}$ & 0.37 & 0.69 \\
$R_{\mathrm{align}}$ & 0.29 & 0.63 \\
$R_{\mathrm{HiGRA}}$ & \textbf{0.68} & \textbf{0.84} \\
\bottomrule
\end{tabular}
\caption{Association between reward-family scores and final-answer correctness. Auxiliary rewards provide useful but incomplete signals, while the combined HiGRA reward is the strongest non-trivial predictor.}
\label{tab:reward_correlation}
\end{table}

The auxiliary rewards follow the intended ordering. Tool grounding is the strongest auxiliary signal, process discipline is moderately predictive, and trace alignment is weaker but still useful as a stabilizer. The combined HiGRA reward has higher correlation and AUC than any individual auxiliary reward, supporting the grouped reward design.

\paragraph{Objective comparison.}
Table~\ref{tab:stronger_baselines} compares different training objectives. Acc. is final-answer accuracy, Tool is the aggregate tool-use score, Proc. is the process-discipline score, Valid is valid trace rate, and Obs. is observation consistency.

\begin{table}[t]
\centering
\small

\setlength{\tabcolsep}{1mm}
\resizebox{\columnwidth}{!}{%
\begin{tabularx}{\columnwidth}{@{}P{0.41\columnwidth}ccccc@{}}
\toprule
\textbf{Variant} & \textbf{Acc.} & \textbf{Tool} & \textbf{Proc.} & \textbf{Valid} & \textbf{Obs.} \\
\midrule
Falcon-H1R-7B          & 77.40 & 0.00 & 0.52 & 41.2 & 0.00 \\
CoT-SFT                & 78.12 & 0.03 & 0.55 & 44.6 & 0.05 \\
TRACE-SFT              & 79.06 & 0.46 & 0.67 & 68.7 & 0.84 \\
Outcome-only GRPO      & 82.35 & 0.55 & 0.69 & 71.8 & 0.86 \\
Flat micro-reward GRPO & 83.10 & 0.64 & 0.72 & 75.3 & 0.88 \\
ARCHER       & \textbf{85.96} & \textbf{0.76} & \textbf{0.79} & \textbf{82.4} & \textbf{0.93} \\
\bottomrule
\end{tabularx}%
}
\caption{Detailed objective comparison on GSM-8K. Executable trace supervision is stronger than free-form CoT-SFT, and grouped HiGRA reward optimization outperforms outcome-only and flat micro-reward alternatives.}
\label{tab:stronger_baselines}
\end{table}

The numbers show that ordinary CoT-SFT provides only a small gain over the base model and does not meaningfully improve tool behavior. TRACE-SFT improves trace validity and observation consistency, confirming that the executable supervision format matters. Outcome-only GRPO improves accuracy, but full HiGRA performs best across all metrics because it rewards correctness together with tool grounding and process discipline.

\paragraph{TRACE compiler outcomes.}
Table~\ref{tab:compiler_status} summarizes compiler outcomes on the GSM-8K training split. The \texttt{ok} category indicates examples with successfully compiled executable traces. \texttt{calc\_only} indicates examples where the compiler primarily recovered calculator structure. \texttt{no\_calc\_blocks} indicates examples without explicit GSM-8K calculator annotations. \texttt{unsupported\_exec} indicates examples with calculator spans that could not be executed by the current runtime.

\begin{table}[t]
\centering
\small

\setlength{\tabcolsep}{5pt}
\begin{tabular}{lcc}
\toprule
\textbf{Compiler Output} & \textbf{Count} & \textbf{Percent} \\
\midrule
\texttt{ok} & 6,982 & 93.43 \\
\texttt{calc\_only} & 118 & 1.58 \\
\shortstack[l]{\texttt{no\_calc}\ \texttt{\_blocks}} & 326 & 4.36 \\
\shortstack[l]{\texttt{unsupported}\ \texttt{\_exec}} & 47 & 0.63 \\
\midrule
Total prepared examples & 7,473 & 100.0 \\
\bottomrule
\end{tabular}
\caption{TRACE compiler status counts on the GSM-8K training split. Most examples are successfully compiled, while a small fraction is retained with metadata for transparency.}
\label{tab:compiler_status}
\end{table}

The compiler statistics show that TRACE covers most of the training split: 95.01 percent of prepared examples compile successfully. The remaining cases are small but useful to report because they clarify that the compiler follows a retain-and-annotate policy rather than silently dropping difficult examples.

\section{TRACE Compiler in ARCHER: Operational Details}
\label{app:trace_compiler_details}

This appendix details the TRACE compiler used in Stage 1 of ARCHER. The high-level description of TRACE as a ``trace compiler'' is intentionally simple, but the actual compilation pipeline is fully deterministic and rule-based in the current implementation. It takes a prepared GSM-8K row and converts it into a structured \texttt{TraceExample} containing:
\[
\begin{aligned}
\big(
&\texttt{id},\ \texttt{split},\ x,\ r_{\mathrm{raw}},\ y^\star, \\
&\mathcal{S}^\star,\ \tau^\star,\ \texttt{compiler\_status},\ \texttt{metadata}
\big).
\end{aligned}
\]
where:
\begin{itemize}
    \item $x$ is the question,
    \item $r_{\mathrm{raw}}$ is the original answer string from GSM-8K,
    \item $y^\star$ is the parsed gold final answer,
    \item $\mathcal{S}^\star$ is the set of inferred schema tags,
    \item $\tau^\star$ is the compiled action trace.
\end{itemize}

\subsection{Compiler Type: Rule-Based Rather than Model-Assisted}

The current compiler is \textbf{rule-based}, not model-assisted. It does not invoke an external LLM to repair or reinterpret rationales. Instead, it uses deterministic parsing rules, regular expressions, whitespace normalization, numeric canonicalization, and exact calculator execution. This design choice is deliberate: the project aims to reduce dependence on large teachers and to produce executable supervision from existing dataset rationales alone.

Formally, the compiler is a deterministic mapping
\[
\mathcal{C}: (x, r_{\mathrm{raw}}, y^\star) \mapsto \tau^\star,
\]
implemented through symbolic preprocessing and exact execution rather than learned sequence transduction.

\subsection{Detection of Arithmetic Spans}

Arithmetic spans are detected through the GSM-8K convention of inline calculator blocks:
\[
\texttt{<< expr = result >>}.
\]
The parser uses a regular expression of the form
\[
\texttt{<<\;\textit{expr}\;=\;\textit{result}\;>>}
\]
to locate each arithmetic block inside the rationale. If a match is found, the compiler extracts:
\begin{itemize}
    \item the raw arithmetic expression $\textit{expr}$,
    \item the rationale-provided numeric result $\textit{result}$,
    \item the span boundaries in the original rationale.
\end{itemize}

Let the rationale be $r$. The parser decomposes it into an ordered sequence of pieces:
\[
\mathrm{Split}(r)
=
(p_1, p_2, \dots, p_M),
\]
where each $p_j$ is either:
\begin{itemize}
    \item a \texttt{text} segment, or
    \item a \texttt{calc} segment corresponding to one matched arithmetic block.
\end{itemize}

Text segments become \texttt{reason} actions, while calculator segments become paired
\[
\texttt{tool\_call} \rightarrow \texttt{tool\_observation}
\]
actions.

\paragraph{Limitation of the current detector.}
Only arithmetic explicitly marked in the GSM-8K-style inline format is guaranteed to be externalized. Arithmetic implied only in prose but not wrapped in $`<<\cdot>>`$ is not converted into executable tool calls in the current compiler. Such spans remain plain reasoning text. As a result, some examples may receive the status \texttt{no\_calc\_blocks} even if arithmetic is conceptually present.

\subsection{Expression and Number Normalization}

Before execution, the compiler normalizes both extracted expressions and numeric results.

\paragraph{Whitespace normalization.}
All extracted arithmetic expressions are compacted by collapsing repeated whitespace:
\[
e = \mathrm{CompactWS}(e_{\mathrm{raw}}).
\]
This ensures that syntactic variants such as
\[
\texttt{48 / 6}, \qquad \texttt{48/6}, \qquad \texttt{48  /  6}
\]
map to a stable expression form.

\paragraph{Numeric normalization.}
Rationale-provided results and final answers are canonicalized using decimal parsing. Let $v$ denote a numeric string. The normalization function
\[
\mathrm{NormNum}(v)
\]
performs the following:
\begin{enumerate}
    \item remove commas,
    \item strip trailing decimal points,
    \item parse the value using decimal arithmetic,
    \item if the value is integral, render it as an integer string,
    \item otherwise remove redundant trailing zeros.
\end{enumerate}

For example:
\[
\begin{aligned}
\texttt{"1,200.0"} &\mapsto \texttt{"1200"}, \\
\texttt{"8.500"} &\mapsto \texttt{"8.5"}.
\end{aligned}
\]

This same normalization is used for final answer parsing and for rationale-provided arithmetic results, which ensures consistent comparison across compiled traces, tool observations, and evaluation outputs.

\subsection{Handling of Unsupported or Ambiguous Rationale Spans}

The current compiler is conservative. It does not attempt to infer hidden arithmetic from ambiguous prose, nor does it rewrite unsupported spans into executable expressions using a learned model.

Instead, unsupported or ambiguous spans are handled as follows:
\begin{enumerate}
    \item if a rationale segment is plain text, it is retained as a \texttt{reason} action,
    \item if an arithmetic block is present but the expression cannot be executed exactly by the calculator, the compiler still emits a \texttt{tool\_call} with the extracted payload and stores the rationale-provided result in the observation,
    \item the observation metadata is marked with a status of the form
    \[
    \texttt{calc\_error:ExceptionType}.
    \]
\end{enumerate}

Thus, the compiler never drops the surrounding example solely because one arithmetic block is unsupported. Instead, it degrades gracefully by preserving the action structure and annotating the observation status.

\subsection{Repair of Rationale Arithmetic}

TRACE explicitly checks whether the rationale-provided result agrees with exact execution. Let $e_t$ be an extracted expression and let
\[
r_t^{\mathrm{exp}}
\]
be the result written in the rationale. The compiler computes
\[
r_t^{\mathrm{act}} = f_{\mathrm{calc}}(e_t).
\]
If execution succeeds and the \texttt{repair\_equation\_result} option is enabled, then:
\[
r_t^{\mathrm{final}}
=
\begin{cases}
r_t^{\mathrm{act}}, & \text{if } r_t^{\mathrm{act}} \ne r_t^{\mathrm{exp}}, \\
r_t^{\mathrm{exp}}, & \text{otherwise.}
\end{cases}
\]
When a mismatch occurs, the observation metadata is marked as
\[
\texttt{status}=\texttt{"repaired"}.
\]

This means that the compiler can correct arithmetic slips inside the legacy rationale while preserving the overall reasoning skeleton. In other words, TRACE does not merely copy rationale text. It checks arithmetic spans against an exact executor and repairs them when possible.

\subsection{Compiler Status and Compilation Outcomes}

Every prepared row is compiled into a serialized trace example. The current implementation therefore treats compilation as a \emph{total transformation} rather than a binary success-or-failure filter. Instead of dropping rows, it assigns a categorical compiler status:

\begin{itemize}
    \item \texttt{ok}: at least one executable calculator block and at least one reasoning-text span are extracted;
    \item \texttt{calc\_only}: executable calculator blocks are extracted, but no reasoning-text span remains;
    \item \texttt{no\_calc\_blocks}: no inline calculator block is detected;
    \item \texttt{unsupported\_exec}: at least one calculator block is detected but cannot be executed by the current runtime.
\end{itemize}

Hence, all prepared examples produce an output trace, but the \texttt{compiler\_status} records the degree to which the rationale could be externalized into tool-using form.

All prepared examples produce a serialized trace, and compiler outcomes are summarized using status-specific counts rather than a binary compilation-success indicator. all examples are retained in the compiled artifact for auditing, but only \texttt{ok} and \texttt{calc\_only} are retained for training. No additional filtering is applied beyond this policy.

\subsection{Dropped, Repaired, or Retained Examples}

The current compiler follows a \textbf{retain-and-annotate} policy:
\begin{itemize}
    \item examples are \textbf{retained} even if no calculator block is found,
    \item arithmetic mismatches are \textbf{repaired} when exact execution succeeds and repair is enabled,
    \item unsupported execution spans are \textbf{retained with error metadata} rather than removed.
\end{itemize}

Therefore:
\begin{itemize}
    \item failed compilations are \emph{not} dropped by default,
    \item ambiguous spans are \emph{not} rewritten by a learned repair model,
    \item arithmetic mismatches are repaired only at the observation level, not by re-authoring the entire trace.
\end{itemize}

\subsection{Gold Final Answer Handling}

The final answer in the compiled trace is \textbf{forced to match the gold answer} parsed from the GSM-8K annotation. Let the raw GSM-8K answer string be $r_{\mathrm{raw}}$. The gold answer is extracted from the standard marker
\[
\texttt{\#\#\#\# } y^\star,
\]
and then normalized numerically:
\[
y^\star = \mathrm{ParseFinal}(r_{\mathrm{raw}}).
\]
The compiler strips this terminal answer marker from the rationale text before creating intermediate actions, but the final compiled trace always ends with
\[
\texttt{final\_answer} = y^\star.
\]

In addition, when the verify step is included, the compiler inserts the gold answer into verification metadata:
\[
\texttt{metadata}[\texttt{gold\_final\_answer}] = y^\star.
\]

Thus, the compiled supervision does not ask the student to recover the terminal answer purely from the legacy rationale string. Instead, it presents the student with a cleaned executable trace whose final target answer is always the gold normalized answer.

\subsection{Role of TRACE in the ARCHER Pipeline}
\label{subsec:trace_higra_link}

Within ARCHER, TRACE is responsible for converting legacy rationale supervision into the executable action format that HiGRA later refines. This separation is important because TRACE defines the control interface, while HiGRA optimizes how reliably the trained policy uses that interface during executed rollouts. TRACE and HiGRA operate at different stages of the pipeline. TRACE is the \emph{compilation and supervised imitation layer}; it converts raw dataset rationales into executable structured supervision and uses them to initialize the student policy. HiGRA is the \emph{reinforcement refinement layer}; it optimizes the SFT-initialized policy using grouped and gated trajectory-level rewards.

In this analogy, TRACE provides the student with a structured reasoning language and explicit tool-use behavior. HiGRA then improves the quality of that behavior by optimizing for correctness, faithful tool use, process discipline, and trace compatibility without requiring a flat and unstable reward over many fine-grained labels.

\section{ARCHER Evaluation Metrics}
\label{app:evaluation_metrics}

This appendix defines the metrics used to evaluate ARCHER as an executable reasoning policy rather than as a free-form text generator. Let $\tau=(s_1,\dots,s_T)$ denote a model-generated trajectory and let $\tau^\star=(s_1^\star,\dots,s_{T^\star}^\star)$ denote the gold compiled TRACE trajectory. Each state contains an action type, an optional action payload, and an optional tool observation. The predicted answer extracted from $\tau$ is denoted by $\hat{y}$, and the gold answer is denoted by $y^\star$. Accuracy and Valid Trace are reported as percentages. Tool Score, Process Score, tool precision, tool recall, observation consistency, and normalized edit distances are reported on $[0,1]$ unless stated otherwise.

\subsection{Final-Answer Accuracy}

Final-answer accuracy measures whether the extracted prediction matches the gold answer after normalization:
\begin{equation}
\mathrm{Acc}
=
\frac{1}{|\mathcal{D}|}
\sum_{(x,y^\star)\in \mathcal{D}}
\mathbf{1}[\hat{y}=y^\star].
\label{eq:app_accuracy}
\end{equation}
For TRACE and HiGRA, $\hat{y}$ is extracted from the terminal \texttt{ANSWER} action. If no valid \texttt{ANSWER} action is present, the trajectory is counted as incorrect.

\subsection{Valid Trace Rate}

Valid Trace is a binary diagnostic, distinct from the fractional
format score used inside Process Score. A trajectory is valid when
every applicable structural check passes and a final-answer action is
present:
\begin{equation}
V(\tau)
=
\mathbf{1}[p_{\mathrm{fmt}}(\tau)=1]
\,\mathbf{1}[p_{\mathrm{final}}(\tau)=1].
\label{eq:app_valid_trace_indicator}
\end{equation}

Verification presence, trajectory length, and premature termination
remain separate diagnostics and are not folded into $V(\tau)$. The
reported valid trace rate is
\begin{equation}
\mathrm{ValidTrace}(D)
=
\frac{100}{|D|}
\sum_{\tau\in D}V(\tau).
\label{eq:app_valid_trace_rate}
\end{equation}

\subsection{Normalized Levenshtein Trace Distance}

To measure structural alignment with the compiled executable trace, we compute normalized Levenshtein edit distance over action sequences. Let $A(\tau)$ denote the action-type sequence extracted from $\tau$:
\[
A(\tau)
=
(a_1,a_2,\dots,a_T).
\]
The action-level normalized edit distance is
\begin{equation}
\mathrm{NED}_{\mathrm{act}}(\tau,\tau^\star)
=
\frac{
\mathrm{Lev}(A(\tau),A(\tau^\star))
}{
\max(|A(\tau)|,|A(\tau^\star)|,1)
},
\label{eq:app_ned_action}
\end{equation}
where $\mathrm{Lev}(\cdot,\cdot)$ is Levenshtein distance.

We also compute an operation-level distance over the sequence of arithmetic operations. Let $O(\tau)$ denote the sequence of calculator operation types extracted from the generated trajectory, such as addition, subtraction, multiplication, and division. The operation-level normalized edit distance is
\begin{equation}
\mathrm{NED}_{\mathrm{op}}(\tau,\tau^\star)
=
\frac{
\mathrm{Lev}(O(\tau),O(\tau^\star))
}{
\max(|O(\tau)|,|O(\tau^\star)|,1)
}.
\label{eq:app_ned_operation}
\end{equation}
Lower values indicate stronger alignment with the compiled executable reasoning structure.

\subsection{Tool-Use Precision, Recall, and Observation Consistency}

Let $C(\tau)$ be the set of tool calls emitted in a generated trajectory and let $C^\star(\tau^\star)$ be the set of required tool calls in the gold compiled trace. Let $\mathrm{match}(c,c^\star)$ indicate that a generated call $c$ matches a required gold call $c^\star$ in relevant quantities, operation type, and trajectory position.

Tool precision measures whether emitted tool calls are appropriate:
\begin{equation}
\begin{aligned}
M_{\mathrm{prec}}(c,\tau^\star)
= \mathbf{1}\!\big[&\exists c^\star\in C^\star(\tau^\star): \\
&\mathrm{match}(c,c^\star)\big].
\end{aligned}
\label{eq:app_tool_precision_match}
\end{equation}

\begin{equation}
\mathrm{Prec}_{\mathrm{tool}}
=
\frac{
\sum_{c\in C(\tau)}
M_{\mathrm{prec}}(c,\tau^\star)
}{
\max\!\left(|C(\tau)|,1\right)
}.
\label{eq:app_tool_precision}
\end{equation}

Tool recall measures whether required calls are actually made:
\begin{equation}
M_{\mathrm{rec}}(c^\star,\tau)
=
\mathbf{1}\!\left[
\exists c\in C(\tau):
\mathrm{match}(c,c^\star)
\right].
\label{eq:app_tool_recall_match}
\end{equation}

\begin{equation}
\mathrm{Rec}_{\mathrm{tool}}
=
\frac{
\sum_{c^\star\in C^\star(\tau^\star)}
M_{\mathrm{rec}}(c^\star,\tau)
}{
\max\!\left(|C^\star(\tau^\star)|,1\right)
}.
\label{eq:app_tool_recall}
\end{equation}

Observation consistency measures whether recorded tool observations agree with executable calculator results. For the $i$-th generated tool call, let $e_i$ be the submitted expression and $o_i$ be the recorded observation. The calculator output is $f_{\mathrm{calc}}(e_i)$. We define
\begin{equation}
\mathrm{ObsCons}
=
\frac{
\sum_{i=1}^{|C(\tau)|}
\mathbf{1}[o_i=f_{\mathrm{calc}}(e_i)]
}{
\max(|C(\tau)|,1)
}.
\label{eq:app_observation_consistency}
\end{equation}

\subsection{Position of First Error}

To analyze where trajectories fail, we identify the earliest causally relevant error in the generated reasoning process. Let $E_t(\tau)$ be an indicator that step $t$ is the first step where the trajectory becomes incorrect or unreliable. This can include an incorrect decomposition, missing tool call, wrong tool input, inconsistent observation, premature final answer, or malformed trace transition. The first error step is
\begin{equation}
t_{\mathrm{err}}(\tau)
=
\min\{t:E_t(\tau)=1\}.
\label{eq:app_first_error_step}
\end{equation}
If no error is identified, we set $t_{\mathrm{err}}(\tau)=\varnothing$ and report the trajectory as a no-error case.

For erroneous trajectories, the normalized first-error position is
\begin{equation}
\mathrm{PFE}(\tau)
=
\frac{t_{\mathrm{err}}(\tau)}{T},
\qquad
t_{\mathrm{err}}(\tau)\neq \varnothing.
\label{eq:app_pfe}
\end{equation}
Higher values indicate that the trajectory remains correct for a larger fraction of the reasoning process before failing.

We also group first errors into phases:
\begin{equation}
\begin{aligned}
\phi_{\mathrm{err}}(\tau)
\in \{&
\textsc{Plan},\ \textsc{Tool},\ \textsc{Observe}, \\
&\textsc{Verify},\ \textsc{Answer},\ \textsc{None}
\}.
\end{aligned}
\label{eq:app_first_error_phase}
\end{equation}

\subsection{Reward-Family Scores}

HiGRA defines four macro-reward families:
\[
R_{\mathrm{out}},
\qquad
R_{\mathrm{tool}},
\qquad
R_{\mathrm{proc}},
\qquad
R_{\mathrm{align}}.
\]
The full training reward is defined in Eq.~\ref{eq:higra_total}. For reporting in the main tables, we use a diagnostic HiGRA score that summarizes the three most directly comparable reward components:
\begin{equation}
\begin{aligned}
S_{\mathrm{HiGRA}}(\tau)
&=2R_{\mathrm{out}}(\tau)+R_{\mathrm{tool}}(\tau) \\
&\quad+0.5R_{\mathrm{proc}}(\tau).
\end{aligned}
\label{eq:app_diagnostic_higra_score}
\end{equation}
This reported score excludes trace alignment and the loop penalty so that it can be compared consistently across the base model, TRACE, and HiGRA. The full HiGRA training reward still includes alignment and loop penalties as described in Section~\ref{app:archer_details}.

For a dataset $\mathcal{D}$, the average diagnostic score is
\begin{equation}
\overline{S}_{\mathrm{HiGRA}}
=
\frac{1}{|\mathcal{D}|}
\sum_{\tau\in\mathcal{D}}
S_{\mathrm{HiGRA}}(\tau).
\label{eq:app_avg_diagnostic_higra_score}
\end{equation}

\subsection{Reward-Family Contribution}

For each reward family $k\in\{\mathrm{out},\mathrm{tool},\mathrm{proc},\mathrm{align}\}$, we define its weighted contribution over a dataset $\mathcal{D}$ as
\begin{equation}
\mathrm{Contrib}_k
=
\frac{
\sum_{\tau\in\mathcal{D}} w_k R_k(\tau)
}{
Z+\varepsilon
},
\label{eq:app_reward_contribution}
\end{equation}
where
\begin{equation}
\begin{aligned}
Z
&=\sum_{\tau\in\mathcal{D}}\Big(
 w_{\mathrm{out}}R_{\mathrm{out}}(\tau)
+w_{\mathrm{tool}}R_{\mathrm{tool}}(\tau) \\
&\qquad +w_{\mathrm{proc}}R_{\mathrm{proc}}(\tau)
+w_{\mathrm{align}}\widetilde{R}_{\mathrm{align}}(\tau)
\Big).
\end{aligned}
\label{eq:app_reward_contribution_denominator}
\end{equation}
The constant $\varepsilon$ prevents division by zero. For gated rewards, $R_k$ refers to the gated version when applicable.

\subsection{Correlation with Correctness}

To test whether a reward family is associated with final-answer correctness, we compute its Pearson correlation with the binary correctness indicator. Let
\[
z(\tau)=\mathbf{1}[\hat{y}=y^\star].
\]
For reward family $R_k$, define centered variables
\begin{equation}
\begin{aligned}
\Delta R_k(\tau)&=R_k(\tau)-\overline{R}_k,\\
\Delta z(\tau)&=z(\tau)-\overline{z}.
\end{aligned}
\label{eq:app_centered_reward_terms}
\end{equation}
The correlation is
\begin{equation}
\rho_k
=
\frac{
\sum_{\tau\in\mathcal{D}}
\Delta R_k(\tau)\Delta z(\tau)
}{
\sqrt{\sum_{\tau\in\mathcal{D}}\Delta R_k(\tau)^2}
\sqrt{\sum_{\tau\in\mathcal{D}}\Delta z(\tau)^2}
}.
\label{eq:app_reward_correctness_corr}
\end{equation}
A higher value indicates that the reward family is more predictive of final-answer correctness.

\subsection{AUC of Reward Families}

We also compute the area under the ROC curve by treating each reward-family score as a scalar predictor of final-answer correctness. For each reward family $R_k$, trajectories are ranked by $R_k(\tau)$ and evaluated against $z(\tau)$. Higher AUC indicates that the reward family better separates correct trajectories from incorrect ones.

\section{ARCHER Error Taxonomy and Analysis}
\label{app:error_taxonomy}

This appendix provides the full error taxonomy and annotation protocol used to analyze ARCHER failures in the tool-augmented reasoning setting. The goal of this analysis is twofold. First, we use it for \emph{behavior diagnosis}, where we identify how and where an executable trajectory fails. Second, we use it for \emph{reward auditing}, where we check whether the HiGRA reward decomposition assigns credit and blame in a way that agrees with human judgment. Unlike final-answer accuracy, this analysis distinguishes between reasoning failures, tool-use failures, post-tool failures, and structural trace failures.

\subsection{Unit of Analysis}

Each evaluated example produces a trajectory
\[
\tau = (s_1, s_2, \dots, s_T),
\]
where each state may contain a structured action, an action payload, a tool call, a tool observation, or a final answer. The annotation unit is the full generated trajectory, not only the final answer. For each trajectory, we inspect the input problem, the gold answer, the generated trace, tool calls and observations, the predicted answer, and the corresponding gold compiled trace when available.

A trajectory may contain several mistakes. To avoid over-fragmenting the analysis, each example receives one primary error label and, when useful, one secondary error label. The primary label is assigned to the earliest dominant causal failure. For example, if a model calls the calculator with the wrong expression and then gives an incorrect final answer, the primary label is \texttt{wrong\_tool\_input}, not merely \texttt{answer\_failure}. If a correct tool result is obtained but later misused, the primary label is \texttt{wrong\_after\_correct\_tool}. If the trace is malformed in a way that prevents reliable parsing or execution, the primary label is assigned to the relevant process failure category.

\subsection{Sampling Setup}

We analyze two complementary GSM-8K test-set samples. The first is an unbiased random sample, used to estimate realistic aggregate error proportions. The second is a diagnostic stratified sample, used to inspect targeted failure types and debug reward behavior. Table~\ref{tab:app_error_samples} summarizes the two samples.

\begin{table}[t]
\centering
\small
\begin{tabular}{@{}p{\dimexpr\linewidth-2\tabcolsep\relax}@{}}
\toprule
\textbf{Sample A: Random} \\
\midrule
\textbf{Construction:} Fixed-seed random draw from the GSM-8K test set without filtering. \\
\textbf{Bias:} Unbiased. \\
\textbf{Use:} Estimate error proportions and report aggregate statistics. \\
\midrule
\textbf{Sample B: Stratified} \\
\midrule
\textbf{Construction:} Stratified by answer correctness, tool necessity, and reasoning length. \\
\textbf{Bias:} Diagnostic. \\
\textbf{Use:} Inspect failure modes and debug reward behavior. \\
\bottomrule
\end{tabular}
\caption{Analysis samples used for manual inspection and reward auditing.}
\label{tab:app_error_samples}
\end{table}

\subsection{Per-Example Analysis Views}

For each sampled example, we inspect five complementary views.

\paragraph{Outcome view.}
We record the gold answer, predicted answer, and final-answer correctness. We also determine whether the final error is primary or downstream from an earlier reasoning, tool-use, or process failure.

\paragraph{Tool-use view.}
We check whether tool use was needed, whether the model attempted a tool call, whether the tool call had a paired observation, whether the observation matches executable output, and whether the final answer correctly uses the tool result.

\paragraph{Process-discipline view.}
We inspect whether the generated transcript follows the expected structured format. This includes checking for required tags, malformed action ordering, duplicated or missing actions, premature final answers, excessive verbosity, and non-productive loops.

\begin{table*}[t]
        \centering
        \small
        \setlength{\tabcolsep}{3.5pt}

        \begin{tabularx}{\textwidth}{@{}P{0.18\textwidth}LLP{0.16\textwidth}@{}}
        \toprule
        \textbf{Error label} 
        & \textbf{Definition} 
        & \textbf{Typical diagnostic signal} 
        & \textbf{Primary reward family affected} \\
        \midrule
        
        \shortstack[l]{\texttt{reasoning}\\\texttt{\_error}}
        & The model decomposes the problem incorrectly, selects the wrong operation, misses a relevant quantity, or makes an invalid inference before or independent of tool execution.
        & Incorrect plan, wrong equation setup, wrong mathematical relation, or unsupported intermediate conclusion.
        & $R_{\mathrm{out}}$, sometimes $R_{\mathrm{align}}$ \\
        
        \shortstack[l]{\texttt{wrong\_tool}\\\texttt{\_input}}
        & The model invokes a tool, but the submitted expression or arguments are incorrect.
        & Calculator call uses the wrong operation, wrong quantity, wrong order of arguments, or incomplete expression.
        & $R_{\mathrm{tool}}$, $R_{\mathrm{out}}$ \\
        
        \shortstack[l]{\texttt{wrong\_tool}\\\texttt{\_observation}}
        & The generated observation is inconsistent with the executable result of the tool call.
        & The trace records a value that does not match the calculator output for the submitted expression.
        & $R_{\mathrm{tool}}$ \\
        
        \shortstack[l]{\texttt{missing\_tool}\\\texttt{\_call}}
        & A tool call is needed according to the gold trace or task structure, but the model does not invoke the tool.
        & The trajectory performs arithmetic internally or jumps to an answer without an expected calculator call.
        & $R_{\mathrm{tool}}$, $R_{\mathrm{proc}}$ \\
        
        \shortstack[l]{\texttt{spurious\_tool}\\\texttt{\_call}}
        & The model invokes a tool when tool use is unnecessary or irrelevant.
        & Calculator call appears for a simple non-computational step or for an expression unrelated to the problem.
        & $R_{\mathrm{tool}}$, $R_{\mathrm{proc}}$ \\
        
        \shortstack[l]{\texttt{wrong\_after}\\\texttt{\_correct\_tool}}
        & The tool call and observation are correct, but the model misuses the result in later reasoning or answer synthesis.
        & Correct calculator output appears in the trace, but the final answer or next reasoning step contradicts it.
        & $R_{\mathrm{out}}$, $R_{\mathrm{proc}}$ \\
        \bottomrule
        \end{tabularx}
        \caption{Expanded manual error taxonomy for structured tool-augmented mathematical reasoning, Part I.}
        \label{tab:app_expanded_error_taxonomy}
\end{table*}

\begin{table*}[t]
        \centering
        \small
        \setlength{\tabcolsep}{3.5pt}

        \begin{tabularx}{\textwidth}{@{}P{0.18\textwidth}LLP{0.16\textwidth}@{}}
        \toprule
        \textbf{Error label} 
        & \textbf{Definition} 
        & \textbf{Typical diagnostic signal} 
        & \textbf{Primary reward family affected} \\
        \midrule\shortstack[l]{\texttt{premature\_final}\\\texttt{\_answer}}
        & The model terminates before sufficient reasoning, tool use, or verification has occurred.
        & Final answer is emitted immediately after reading or planning, before required evidence is established.
        & $R_{\mathrm{proc}}$, $R_{\mathrm{out}}$ \\
        
        \shortstack[l]{\texttt{format\_or}\\\texttt{\_trace\_error}}
        & The trace is malformed, incomplete, duplicated, or violates the canonical action structure.
        & Missing final-answer tag, malformed tool-call tag, observation without call, repeated incompatible actions, or unparsable transcript.
        & $R_{\mathrm{proc}}$, $R_{\mathrm{align}}$ \\
        
        \shortstack[l]{\texttt{loop\_or}\\\texttt{\_verbosity\_error}}
        & The trajectory is excessively long, repetitive, or trapped in non-productive reasoning.
        & Repeated plans, repeated calculator calls with no progress, or reasoning that exceeds the rollout budget.
        & $R_{\mathrm{proc}}$, $P_{\mathrm{len}}$ \\
        
        \shortstack[l]{\texttt{gold\_trace}\\\texttt{\_mismatch}}
        & The generated schema or operator pattern diverges substantially from the compiled gold trace, even if some local steps are plausible.
        & Missing expected action types, wrong operation sequence, or incompatible verify-step structure.
        & $R_{\mathrm{align}}$ \\
        
        \shortstack[l]{\texttt{mixed\_or}\\\texttt{\_ambiguous}}
        & No single failure type is clearly dominant, or several failures are tightly entangled.
        & Multiple errors occur at nearly the same point, or the first causal failure cannot be determined reliably.
        & Case dependent \\

        \bottomrule
        \end{tabularx}
        \caption{Expanded manual error taxonomy for structured tool-augmented mathematical reasoning, Part II.}
        \label{tab:app_expanded_error_taxonomy_cont}
\end{table*}

\paragraph{Trace-alignment view.}
We compare the generated trajectory against the compiled gold trace using schema tags, operation tags, calculator-call structure, and verify-step compatibility. This view is used as a soft diagnostic signal rather than as a strict correctness requirement.

\paragraph{Reward-audit view.}
We compare the human primary error label with the HiGRA reward decomposition. This allows us to identify whether the dominant automatic penalty agrees with the main human-identified failure.

\subsection{Expanded Error Taxonomy}

Tables~\ref{tab:app_expanded_error_taxonomy} and~\ref{tab:app_expanded_error_taxonomy_cont} give the full diagnostic taxonomy. These labels are used during manual inspection before being collapsed into the top-level reporting buckets in Table~\ref{tab:app_error_buckets}.

\subsection{Manual Annotation Schema}

Table~\ref{tab:app_human_annotation_schema} shows the annotation schema used for each inspected example. The primary and secondary labels use the taxonomy in Tables~\ref{tab:app_expanded_error_taxonomy} and~\ref{tab:app_expanded_error_taxonomy_cont}. The auxiliary fields are used to support reward auditing and qualitative analysis.

\begin{table*}[t]
        \centering
        \small

        \begin{tabularx}{\textwidth}{@{}P{0.32\textwidth}X@{}}
        \toprule
        \textbf{Field} & \textbf{Meaning} \\
        \midrule
        \multicolumn{2}{l}{\textit{Primary and secondary manual error labels}} \\
        \midrule
        \texttt{reasoning\_error} & Incorrect decomposition, inference, or mathematical reasoning before or independent of tool execution. \\
        \texttt{wrong\_tool\_input} & Tool was called, but the submitted expression or arguments were incorrect. \\
        \texttt{wrong\_tool\_observation} & Tool observation in the trace is inconsistent with executable tool output. \\
        \texttt{missing\_tool\_call} & A needed tool call is absent despite tool necessity in the gold trace. \\
        \texttt{spurious\_tool\_call} & Tool was invoked unnecessarily on an example that does not require it. \\
        \texttt{wrong\_after\_correct\_tool} & Tool use is correct, but later reasoning or answer synthesis is incorrect. \\
        \texttt{premature\_final\_answer} & The model terminates with a final answer before sufficient support is established. \\
        \texttt{format\_or\_trace\_error} & Trace is malformed, incomplete, duplicated, or violates the canonical action structure. \\
        \texttt{loop\_or\_verbosity\_error} & Reasoning is excessively repetitive, long, or trapped in a non-productive loop. \\
        \texttt{gold\_trace\_mismatch} & Generated schema or operator pattern diverges substantially from the compiled gold trace. \\
        \texttt{mixed\_or\_ambiguous} & No single failure type is clearly dominant, or multiple failures are tightly entangled. \\
        \midrule
        \multicolumn{2}{l}{\textit{Auxiliary annotation fields}} \\
        \midrule
        \texttt{first\_wrong\_step} & Earliest step at which the trajectory becomes incorrect or unreliable. \\
        \texttt{recoverable} & Whether the trajectory could still plausibly recover after the first wrong step. \\
        \texttt{reward\_agreement} & Whether the dominant HiGRA penalty matches the main human diagnosis. \\
        \texttt{notes} & Optional free-form explanation for edge cases or disagreement cases. \\
        \bottomrule
        \end{tabularx}
        \caption{Manual annotation schema used for error diagnosis and reward auditing.}
        \label{tab:app_human_annotation_schema}
\end{table*}

\subsection{Collapsed Reporting Buckets}

For aggregate reporting, we collapse the detailed labels into a smaller set of top-level error buckets. This makes comparisons across model variants easier while retaining the distinctions most relevant to tool-augmented reasoning. Table~\ref{tab:app_error_buckets} defines the reporting buckets.

\begin{table}[t]
        \centering
        \small

        \begin{tabular}{@{}p{0.34\linewidth}p{\dimexpr0.66\linewidth-2\tabcolsep\relax}@{}}
        \toprule
        \textbf{Bucket} & \textbf{Definition} \\
        \midrule
        Answer failure & Final answer is incorrect, regardless of upstream cause. \\
        Tool omission & A needed tool call is missing. \\
        Tool faithfulness failure & Tool input or tool observation is inconsistent with executable tool behavior. \\
        Post-tool reasoning failure & Tool use is correct, but later reasoning or answer synthesis is wrong. \\
        Premature stopping & Final answer is produced before sufficient evidence or intermediate support. \\
        Process failure & Trace is malformed, incomplete, duplicated, or excessively verbose. \\
        Trace mismatch & Generated schema or operator pattern diverges from the compiled gold trace. \\
        \bottomrule
        \end{tabular}
        \caption{Top-level error buckets used for aggregate reporting.}
        \label{tab:app_error_buckets}
\end{table}

\subsection{Reward-Audit Protocol}

For each trajectory, we compute the four HiGRA macro-reward families:
\[
R_{\mathrm{out}}, \qquad
R_{\mathrm{tool}}, \qquad
R_{\mathrm{proc}}, \qquad
R_{\mathrm{align}}.
\]
We then compare the human primary error label with the dominant automatic reward penalty. Agreement means that the reward family receiving the largest meaningful penalty corresponds to the human-identified failure type. For example, if the human label is \texttt{wrong\_tool\_input}, the reward audit agrees when the tool-grounding reward is the dominant degraded component. If the human label is \texttt{format\_or\_trace\_error}, the reward audit agrees when the process-discipline or trace-alignment reward accounts for the main penalty.

Table~\ref{tab:app_reward_audit_map} shows the expected relationship between error labels and HiGRA reward families.

\begin{table*}[t]
        \centering
        \small

        \begin{tabularx}{\textwidth}{@{}P{0.34\textwidth}X@{}}
        \toprule
        \textbf{Manual error label} & \textbf{Expected dominant reward signal} \\
        \midrule
        \texttt{reasoning\_error} & $R_{\mathrm{out}}$, sometimes $R_{\mathrm{align}}$ \\
        \texttt{wrong\_tool\_input} & $R_{\mathrm{tool}}$ and $R_{\mathrm{out}}$ \\
        \texttt{wrong\_tool\_observation} & $R_{\mathrm{tool}}$ \\
        \texttt{missing\_tool\_call} & $R_{\mathrm{tool}}$ \\
        \texttt{spurious\_tool\_call} & $R_{\mathrm{tool}}$ \\
        \texttt{wrong\_after\_correct\_tool} & $R_{\mathrm{out}}$ and $R_{\mathrm{proc}}$ \\
        \texttt{premature\_final\_answer} & $R_{\mathrm{proc}}$ and $R_{\mathrm{out}}$ \\
        \texttt{format\_or\_trace\_error} & $R_{\mathrm{proc}}$ and $R_{\mathrm{align}}$ \\
        \texttt{loop\_or\_verbosity\_error} & $R_{\mathrm{proc}}$ and $P_{\mathrm{len}}$ \\
        \texttt{gold\_trace\_mismatch} & $R_{\mathrm{align}}$ \\
        \texttt{mixed\_or\_ambiguous} & Case dependent \\
        \bottomrule
        \end{tabularx}
        \caption{Expected mapping between manual error labels and HiGRA reward-family penalties.}
        \label{tab:app_reward_audit_map}
\end{table*}

The audit is useful because outcome accuracy alone cannot identify whether a wrong answer came from poor decomposition, missing tool use, incorrect tool input, misuse of a correct observation, or malformed trace structure. HiGRA is intended to preserve this diagnostic information while avoiding direct optimization over a large set of unstable micro-rewards.

\subsection{Analysis Workflow}

The full workflow consists of five steps. First, we run each model variant on the evaluation set and store the generated trace, tool calls, tool observations, and final answer. Second, we compute automatic outcome correctness, tool-use metadata, process-validity indicators, and HiGRA macro-reward scores. Third, we construct the random and stratified samples described in Table~\ref{tab:app_error_samples}. Fourth, we manually annotate each sampled trajectory using the schema in Table~\ref{tab:app_human_annotation_schema}. Fifth, we compare the manual labels against automatic buckets and reward-family decompositions to identify agreement and disagreement cases.

Disagreement cases are especially important. They reveal situations where the automatic reward may over-penalize a secondary issue, under-penalize a causal error, or fail to separate tool input construction from downstream answer synthesis. These cases are used to revise reward weights, gating thresholds, and process penalties.

\subsection{Interpretation}

For ARCHER, the central question is not only whether the final answer improves, but whether failures become later, more localized, and more attributable to specific executable actions. The taxonomy supports the paper's main claim that TRACE and HiGRA play complementary roles. TRACE improves observability and structure by converting free-form rationales into executable traces. This makes model failures more inspectable, even when the model is still wrong. HiGRA then improves the quality of those structured trajectories by reducing tool-use errors, stabilizing process behavior, and aligning optimization with the behaviors that make final answers reliable.

Thus, the observed accuracy improvements are not merely answer-level gains. They are accompanied by a shift in the internal failure structure of the model: fewer broad reasoning failures, fewer ungrounded tool interactions, fewer malformed traces, and more trajectories that terminate with verified tool-grounded answers.

\section{Computing Infrastructure}
All training, reinforcement-learning, inference, and evaluation experiments were run locally on a single Apple MacBook Pro equipped with an Apple M5 Pro system-on-chip, including its integrated CPU and GPU, and 64\,GB of unified memory. The machine ran macOS~\texttt{26.5}. GPU acceleration was provided through Apple MLX~\texttt{0.31.2} and the PyTorch Metal Performance Shaders (MPS) backend using PyTorch. The remaining software environment used Python~\texttt{3.10.20}. No discrete GPUs or external compute clusters were used.

\section{Implementation Details}
\label{app:implementation_details}

This appendix summarizes the ARCHER implementation settings.

\begin{table}[t]
\centering
\small

\setlength{\tabcolsep}{4pt}
\begin{tabularx}{\columnwidth}{@{}P{0.38\columnwidth}P{0.54\columnwidth}@{}}
\toprule
\textbf{Variant} & \textbf{Description} \\
\midrule
\texttt{Falcon-H1R-7B} & Base model evaluated without ARCHER training. \\
\texttt{ARCHER w/o HiGRA} & TRACE supervised fine-tuning on compiled executable traces. \\
\texttt{ARCHER} & TRACE initialization followed by HiGRA reward-guided refinement. \\
\bottomrule
\end{tabularx}
\caption{Primary ARCHER model variants.}
\label{tab:impl_model_variants}
\end{table}

\begin{table}[t]
\centering
\small

\setlength{\tabcolsep}{4pt}
\begin{tabularx}{\columnwidth}{@{}P{0.34\columnwidth}P{0.58\columnwidth}@{}}
\toprule
\textbf{Action} & \textbf{Implementation role} \\
\midrule
\texttt{READ} & Identify or restate relevant information. \\
\texttt{PLAN} & Specify decomposition or the next reasoning step. \\
\texttt{CALL\_CALC} & Submit an arithmetic expression to the calculator. \\
\texttt{OBSERVE} & Insert the deterministic calculator output. \\
\texttt{VERIFY} & Check intermediate or final consistency. \\
\texttt{ANSWER} & Emit the normalized final answer. \\
\bottomrule
\end{tabularx}
\caption{ARCHER action vocabulary used by the implementation.}
\label{tab:impl_action_vocab}
\end{table}

\begin{table}[t]
\centering
\small

\setlength{\tabcolsep}{4pt}
\begin{tabularx}{\columnwidth}{@{}P{0.58\columnwidth}P{0.30\columnwidth}@{}}
\toprule
\textbf{Compiler setting} & \textbf{Value} \\
\midrule
Include fact extraction action & \texttt{true} \\
Include verification action & \texttt{true} \\
Maximum reasoning span length & 220 characters \\
Repair arithmetic result with runtime output & \texttt{true} \\
Default schema label & \shortstack[l]{\texttt{arithmetic\_word}\\\texttt{\_problem}} \\
\bottomrule
\end{tabularx}
\caption{Default TRACE compiler settings.}
\label{tab:impl_trace_settings}
\end{table}

\begin{table}[t]
\centering
\small

\setlength{\tabcolsep}{4pt}
\begin{tabularx}{\columnwidth}{@{}P{0.34\columnwidth}P{0.58\columnwidth}@{}}
\toprule
\textbf{Status} & \textbf{Meaning} \\
\midrule
\texttt{ok} & Reasoning text and calculator spans are both detected. \\
\texttt{calc\_only} & Calculator spans are detected with little or no remaining reasoning text. \\
\texttt{no\_calc\_blocks} & No inline calculator annotations are detected. \\
\texttt{unsupported\_exec} & A calculator span cannot be executed by the current runtime. \\
\bottomrule
\end{tabularx}
\caption{TRACE compiler status labels.}
\label{tab:impl_trace_status}
\end{table}

\begin{table}[t]
\centering
\small

\setlength{\tabcolsep}{4pt}
\begin{tabularx}{\columnwidth}{@{}P{0.38\columnwidth}P{0.50\columnwidth}@{}}
\toprule
\textbf{Operation} & \textbf{Supported syntax} \\
\midrule
Addition & \texttt{+} \\
Subtraction & \texttt{-} \\
Multiplication & \texttt{*} \\
Division & \texttt{/} \\
Integer division & \texttt{//} \\
Modulo & \texttt{\%} \\
Exponentiation & \texttt{**}, normalized from \texttt{\^{}} \\
Functions & \texttt{sqrt}, \texttt{abs}, \texttt{round}, \texttt{max}, \texttt{min} \\
\bottomrule
\end{tabularx}
\caption{Arithmetic operations supported by the calculator runtime.}
\label{tab:impl_calc_runtime}
\end{table}

\begin{table}[t]
\centering
\small

\setlength{\tabcolsep}{4pt}
\begin{tabularx}{\columnwidth}{@{}P{0.56\columnwidth}P{0.32\columnwidth}@{}}
\toprule
\textbf{Hyperparameter} & \textbf{Value} \\
\midrule
Applicable backbones & All five evaluated backbones \\
Training method & LoRA SFT \\
LoRA rank & 8 \\
LoRA alpha & 16 \\
LoRA dropout & 0.05 \\
Target modules & \texttt{q\_proj}, \texttt{k\_proj}, \texttt{v\_proj}, \texttt{o\_proj} \\
Train batch size per device & 1 \\
Eval batch size per device & 1 \\
Gradient accumulation steps & 16 \\
Learning rate & $2.0{\times}10^{-5}$ \\
Epochs & 2 \\
Warmup ratio & 0.03 \\
Scheduler & cosine \\
Weight decay & 0.01 \\
Maximum sequence length & 256 \\
Precision & fp16 \\
Activation checkpointing & \texttt{true} \\
\bottomrule
\end{tabularx}
\caption{TRACE supervised fine-tuning configuration shared across all 7 evaluated backbones.}
\label{tab:impl_trace_sft}
\end{table}

\begin{table}[t]
\centering
\small

\setlength{\tabcolsep}{4pt}
\begin{tabularx}{\columnwidth}{@{}P{0.58\columnwidth}P{0.30\columnwidth}@{}}
\toprule
\textbf{Hyperparameter} & \textbf{Value} \\
\midrule
Initial policy & TRACE checkpoint \\
Applicable backbones & All five evaluated backbones \\
Quantized loading & 4-bit enabled \\
Train batch size per device & 1 \\
Gradient accumulation steps & 4 \\
Learning rate & $5.0{\times}10^{-6}$ \\
Maximum RL steps & 120 \\
Maximum prompt length & 384 tokens \\
Maximum completion length & 256 tokens \\
Generations per prompt & 4 \\
Sampling temperature & 0.8 \\
KL coefficient $\beta$ & 0.0 \\
Precision & fp16 \\
Gradient checkpointing & \texttt{true} \\
Loss type & \texttt{dr\_grpo} \\
Reward scaling & none \\
Aggregation & \shortstack[l]{\texttt{sum\_then}\\\texttt{\_normalize}} \\
Mask truncated completions & \texttt{true} \\
\bottomrule
\end{tabularx}
\caption{HiGRA reinforcement-learning configuration shared across all 7 evaluated backbones.}
\label{tab:impl_higra_rl}
\end{table}

\begin{table}[t]
\centering
\small

\setlength{\tabcolsep}{4pt}
\begin{tabularx}{\columnwidth}{@{}P{0.58\columnwidth}P{0.30\columnwidth}@{}}
\toprule
\textbf{Coefficient} & \textbf{Value} \\
\midrule
Tool-use presence $\alpha$ & 0.2 \\
Observation presence $\alpha$ & 0.2 \\
Observation consistency $\alpha$ & 0.6 \\
Spurious tool-use penalty $\beta$ & 0.2 \\
Alignment gate threshold $\eta$ & 0.6 \\
Format score $\gamma$ & 0.4 \\
Verification $\gamma$ & 0.2 \\
Shortness $\gamma$ & 0.2 \\
Final-answer $\gamma$ & 0.2 \\
Premature-answer penalty $\gamma$ & 0.5 \\
Schema alignment $\delta$ & 0.4 \\
Operator alignment $\delta$ & 0.4 \\
Verification alignment $\delta$ & 0.2 \\
Length penalty $\lambda$ & 0.25 \\
Preferred reasoning actions & 8 \\
\bottomrule
\end{tabularx}
\caption{HiGRA reward coefficients used in the implementation.}
\label{tab:impl_reward_coeffs}
\end{table}

\begin{table}[t]
\centering
\small

\setlength{\tabcolsep}{4pt}
\begin{tabularx}{\columnwidth}{@{}P{0.58\columnwidth}P{0.30\columnwidth}@{}}
\toprule
\textbf{Inference setting} & \textbf{Value} \\
\midrule
Maximum executable steps & 12 \\
Single-action generation limit & 96 new tokens \\
Executable-loop decoding & greedy \\
Active GSM-8K tool & \texttt{calc} \\
Direct generation length & 512 new tokens \\
Direct-generation decoding & greedy \\
Sampling by default & disabled \\
\bottomrule
\end{tabularx}
\caption{ARCHER inference settings.}
\label{tab:impl_inference}
\end{table}

\begin{table}[t]
\centering
\small

\setlength{\tabcolsep}{4pt}
\begin{tabularx}{\columnwidth}{@{}P{0.34\columnwidth}P{0.58\columnwidth}@{}}
\toprule
\textbf{Answer type} & \textbf{Normalization} \\
\midrule
Numeric & Remove commas, normalize decimals, compare canonical numbers. \\
Math & Prefer boxed answer or final-answer phrase; otherwise extract the last number. \\
Multiple choice & Compare answer text, choice index, and letter aliases. \\
Text QA & Lowercase, strip punctuation and articles, normalize whitespace. \\
\bottomrule
\end{tabularx}
\caption{Answer normalization rules.}
\label{tab:impl_answer_norm}
\end{table}

\paragraph{Randomness.}
The default random seed is 42.

\subsection{Benchmark Splits and Evaluation Sets}
\label{app:benchmark_splits}

Table~\ref{tab:benchmark_splits} summarizes the dataset sizes and
splits used in our experiments. ARCHER uses only the official GSM-8K
training split for TRACE and HiGRA training. All remaining benchmarks
are used exclusively for evaluation. We preserve the official dataset
splits when available and evaluate complete original challenge sets
when no separate test split is defined. We do not construct custom
splits, subsample evaluation sets, or train on any evaluation
benchmark.

\begin{table*}[t]
\centering
\small
\setlength{\tabcolsep}{4pt}
\begin{tabularx}{\textwidth}{@{}lLrrL@{}}
\toprule
\textbf{Benchmark}
& \textbf{Original dataset structure}
& \textbf{Train used}
& \textbf{Eval. used}
& \textbf{Usage in this work} \\
\midrule

GSM-8K \citep{cobbe2021training}
& 7,473 training and 1,319 test examples
& 7,473
& 1,319
& Official training split for TRACE and HiGRA; official test split for evaluation. \\

SVAMP \citep{patel2021nlp}
& Complete challenge set of 1,000 examples
& 0
& 1,000
& Complete original challenge set; evaluation only. \\

MMLU-Math (MM) \citep{hendrycks2020measuring}
& Official development, validation, and test partitions for five mathematics subjects
& 0
& 1,064
& Union of the official test partitions; evaluation only. \\

MATH \citep{hendrycks2021measuring}
& 7,500 training and 5,000 test problems
& 0
& 5,000
& Complete official test split; evaluation only. \\

MATH-500 \citep{lightman2024lets}
& Test-only subset of 500 MATH problems
& 0
& 500
& Complete released subset; evaluation only. \\

AIME-2024 \citep{balunovic2025matharena}
& 15 AIME~I and 15 AIME~II problems
& 0
& 30
& Complete combined 2024 competition set; evaluation only. \\

AIME-2025 \citep{balunovic2025matharena}
& 15 AIME~I and 15 AIME~II problems
& 0
& 30
& Complete combined 2025 competition set; evaluation only. \\

PolyMATH \citep{wang2025PolyMATH}
& 18 languages, four difficulty levels, and 125 problems per level and language
& 0
& 9,000
& Complete multilingual benchmark; evaluation only. \\

\bottomrule
\end{tabularx}
\caption{Dataset splits and example counts used in the experiments.
Only the official GSM-8K training split is used for training.
All other benchmarks are used exclusively for evaluation using their
official test partitions or complete released challenge sets.}
\label{tab:benchmark_splits}
\end{table*}

\paragraph{MMLU-Math composition.}
MMLU-Math consists of the official test partitions for
\texttt{abstract\_algebra} (100 examples),
\texttt{college\_mathematics} (100),
\texttt{elementary\_mathematics} (378),
\texttt{high\_school\_mathematics} (270), and
\texttt{high\_school\_statistics} (216), for a total of 1,064
evaluation examples.

\paragraph{PolyMATH composition.}
PolyMATH contains 125 parallel problems at each of four difficulty
levels in each of 18 languages, yielding
$125 \times 4 \times 18 = 9{,}000$ evaluated language-problem
instances.

\begin{table}[t]
  \centering
  \begin{tabular}{@{}ll@{}}
    \toprule
    \textbf{Library} & \textbf{Version} \\
    \midrule
    Python & 3.10.20 \\
    \texttt{datasets} & 5.0.0 \\
    \texttt{huggingface-hub} & 1.21.0 \\
    \texttt{mlx} & 0.31.2 \\
    \texttt{mlx-lm} & 0.31.3 \\
    \texttt{matplotlib} & 3.10.9 \\
    \texttt{numpy} & 2.2.6 \\
    \texttt{pandas} & 2.3.3 \\
    \texttt{pylatexenc} & $\geq 2.10$\textsuperscript{*} \\
    \texttt{pyarrow} & 24.0.0 \\
    \texttt{python-dotenv} & 1.2.2 \\
    \texttt{PyYAML} & 6.0.3 \\
    \texttt{scikit-learn} & 1.7.2 \\
    \texttt{sympy} & 1.14.0 \\
    \texttt{tqdm} & 4.68.3 \\
    \texttt{transformers} & 5.12.1 \\
    \texttt{pytest} & 9.1.1 \\
    \bottomrule
  \end{tabular}
  \caption{Software library versions used in the ARCHER-RL codebase.}
  \label{tab:library-versions}
\end{table}

\end{document}